\documentclass[twoside]{article}
\usepackage[accepted]{aistats2016_}

\usepackage{times}
\usepackage{graphicx}
\usepackage{subcaption}

\usepackage{amsmath}
\usepackage{amsthm}
\usepackage{verbatim} 
\usepackage{amsfonts}

\usepackage[numbers]{natbib}

\let\chapter\section  

\usepackage{color}

\DeclareMathOperator*{\argmin}{argmin}
\DeclareMathOperator*{\argmax}{argmax}

\newcommand{\reals}{\mathbb{R}}
\newcommand{\secref}[1]{{Section~\ref{#1}}}

\newcommand{\figref}[1]{{Figure~\ref{#1}}}

\DeclareSymbolFont{symbolsC}{U}{txsyc}{m}{n}
\DeclareMathSymbol{\notniFromTxfonts}{\mathrel}{symbolsC}{61}

\usepackage{verbatim} 
\usepackage[utf8]{inputenc}
\usepackage{amsfonts}
\usepackage{amsmath}
\usepackage{amssymb}
\usepackage{enumerate}
\usepackage{amsthm}
\usepackage{color}
\newcommand{\red}[1]{{\textcolor{red}{#1}}}
\newcommand{\todo}[1]{\red{TODO: {#1}}}

\usepackage[ruled]{algorithm}
\usepackage[noend]{algpseudocode}
\usepackage{esvect}
\algnewcommand\algorithmicinput{\textbf{Input:}}
\algnewcommand\INPUT{\item[\algorithmicinput]}
\algnewcommand\algorithmicoutput{\textbf{Output:}}
\algnewcommand\OUTPUT{\item[\algorithmicoutput]}
\algnewcommand\algorithmicinitialize{\textbf{Initialize:}}
\algnewcommand\INIT{\item[\algorithmicinitialize]}

\usepackage{float}
\newfloat{algorithm}{t}{lop}

\newtheorem*{rep@theorem}{\rep@title}
\newcommand{\newreptheorem}[2]{%
\newenvironment{rep#1}[1]{%
 \def\rep@title{#2 \ref{##1}}%
 \begin{rep@theorem}}%
 {\end{rep@theorem}}}
 \makeatother

\newtheorem{theorem}{Theorem}
\newreptheorem{theorem}{Theorem}
\newtheorem{lemma}{Lemma}
\newreptheorem{lemma}{Lemma}

\newtheorem{corollary}[theorem]{Corollary}
\newreptheorem{corollary}{Corollary}

\begin{document}

%

%

\twocolumn[

\aistatstitle{Fast and Scalable Structural SVM with Slack Rescaling}

\aistatsauthor{ Heejin Choi \And Ofer Meshi \And Nathan Srebro }

\aistatsaddress{ Toyota Technological Institute at Chicago\\ 6045 S. Kenwood Ave.
Chicago, Illinois 60637
USA} ]

%




\begin{abstract}
  We present an efficient method for training slack-rescaled
  structural SVM. Although finding the most violating label in a
  margin-rescaled formulation is often easy since the target function
  decomposes with respect to the structure, this is not the case for a
  slack-rescaled formulation, and finding the most violated label
  might be very difficult.  Our core contribution is an efficient
  method for finding the most-violating-label in a slack-rescaled
  formulation, given an oracle that returns the most-violating-label
  in a (slightly modified) margin-rescaled formulation.  We show that
  our method enables accurate and scalable training for slack-rescaled
  SVMs, reducing runtime by an order of magnitude compared to previous
  approaches to slack-rescaled SVMs.
\end{abstract}



 \section{Introduction}
Many problems in machine learning can be seen as structured output prediction tasks, where one would like to predict a set of labels with rich internal structure \cite{struct_pred_book}.
This general framework has proved useful for a wide range of applications from computer vision, natural language processing, computational biology, and others.
In order to achieve high prediction accuracy, the parameters of structured predictors are learned from training data.
One of the most effective and commonly used approaches for this supervised learning task is \emph{Structural SVM}, a method that generalizes binary SVM to structured outputs \cite{tsochantaridis2004support}.
Since the structured error is non-convex, \citet{tsochantaridis2004support} propose to replace it with a convex surrogate loss function.
They formulate two such surrogates, known as \emph{margin} and \emph{slack rescaling}.

While slack rescaling often produces more accurate predictors, margin
rescaling has been far more popular due to its better computational
requirements.  In particular, both formulations require optimizing
over the output space, but while margin rescaling preserves the
structure of the score and error functions, the slack-rescaling does
not. This results in harder inference problems during training.  To
address this challenge, \citet{sarawagi2008accurate} propose a method
to reduce the problem of slack rescaling to a series of modified
margin rescaling problems.  They show that their method outperforms
margin rescaling in several domains.  However, there are two main
caveats in their approach. First, the optimization is only heuristic,
that is, it is not guaranteed to solve the slack rescaling objective
exactly. Second, their method is specific to the cutting plane
algorithm and does not easily extend to stochastic training algorithms.
More recently, \citet{bauer2014efficient} proposed an elegant dynamic
programming approach to the slack rescaling optimization problem.
However, their formulation is restricted to sequence labeling and
hamming error, and does not apply to more general structures.

In this paper we propose an efficient method for solving the
optimization problem arising from slack rescaling formulation.
Similar to \citet{sarawagi2008accurate} our method reduces finding the
most violated label in slack rescaling to a series of margin rescaling
problems.  However, in contrast to their approach, our approach can be
easily used with training algorithms like stochastic gradient descent
(SGD) \cite{RatliffSubgradient} and block Frank-Wolfe (FW)
\cite{lacoste2013block}, which often scale much better than cutting
plane.  We first propose a very simple approach that minimizes an
upper bound on the slack rescaling objective function, and only
requires access to a margin rescaling oracle.  This formulation is
quite general and can be used with any error function and model
structure, and many training algorithms such as cutting plane, SGD and
FW.  However, this method is not guaranteed to always find the most
violating label.  Indeed, we show that always finding the most
violating label for a slack-rescaled formulation is impossible using
only a margin rescaling oracle.  To address this, we suggest using a
modified oracle, that is typically as easy to implement as margin
rescaling, and present a more sophisticated algorithm which solves
the slack rescaling formulation exactly, and also enjoys good
approximation guarantees after a small number of iterations.  We
demonstrate empirically that our algorithm outperforms existing
baselines on several real-world applications, including hierarchical
and multi-label classification.

\section{Problem Formulation}
In this section we review the basics of structured output prediction
and describe the relevant training objectives.  In structured output
prediction the task is to map data instances $x$ to a set of output
labels $y\in\mathcal{Y}$.  Structured SVMs use a linear discriminant
mapping of the form $y(x;w) = \argmax_{y\in\mathcal{Y}}
w^\top\phi(x,y)$, where $\phi(x,y)\in\reals^d$ is a feature function
relating input-output pairs, and $w\in\reals^d$ is a corresponding
vector of weights.  Our interest is in the supervised learning
setting, where $w$ is learned from training data $\{x_i,y_i\}^n_{i=1}$
by minimizing the empirical risk.  The prediction quality is measured
by an error function $L(y,y_i)\ge 0$ which determines how bad it is to
predict $y$ when the ground-truth is in fact $y_i$.

Since optimizing $L(y,y_i)$ directly is hard due to its complicated dependence on $w$, several alternative formulations minimize a convex upper bound instead.
Structural SVM is an elegant max-margin approach which uses a structured hinge loss surrogate \cite{tsochantaridis2004support,Taskar03}.
Two popular surrogates are margin and slack rescaling.
In particular, denoting the model score by $f(y) = w^\top\phi(x,y)$ (we omit the dependence on $x$ and $w$ to simplify notation), the margin rescaling training objective is given by:
 \begin{align}\label{obj_margin}
        &\min_{w,\xi} \dfrac{C}{2}\|w\|^2_2 +\dfrac{1}{n}\sum_i\xi_i \\
    \text{s.t.} \;\;& f(y_i)-f(y)\ge L(y,y_i)-\xi_i &\forall i,y\ne y_i \nonumber\\
     & \xi_i\ge 0 &\forall i\nonumber
 \end{align}
 where $C$ is the regularization constant.
Similarly, the slack rescaling formulation scales the slack variables by the error term:
 \begin{align}\label{obj_slack}
        &\min_{w,\xi}\ \dfrac{C}{2}\|w\|^2_2 +\dfrac{1}{n}\sum_i\xi_i \\
    \text{s.t.} \;\;& f(y_i)-f(y)\ge 1- \dfrac{\xi_i}{L(y,y_i)} &\forall i,y\ne y_i \nonumber\\
     & \xi_i\ge 0 &\forall i \nonumber
 \end{align}

Intuitively, both formulations seek to find a $w$ which assigns high scores to the ground-truth compared to the other possible labellings. When $y$ is very different than the true $y_i$ ($L$ is large) then the difference in scores should also be larger.
There is, however, an important difference between the two forms. In margin rescaling, high loss can occur for labellings with high error even though they are already classified correctly with a margin. This may divert training from the interesting labellings where the classifier errs, especially when $L$ can take large values, as in the common case of hamming error. In contrast, in slack rescaling labellings that are classified with a margin incur no loss.
Another difference between the two formulations is that the slack rescaling loss is invariant to scaling of the error term, while in margin rescaling such scaling changes the meaning of the features $\phi$.

In many cases it is easier to optimize an unconstrained problem.
In our case it is easy to write \eqref{obj_margin} and \eqref{obj_slack} in an unconstrained form:
{\small \begin{align} 
\label{unconstrained_obj_margin}
        &\min_{w} \tfrac{C}{2}\|w\|^2_2
        +\!\tfrac{1}{n}\!\sum_i\max_{y\in\mathcal{Y}} \left( L(y,y_i)+f(y)-f(y_i)\right) \\
\label{unconstrained_obj_slack}
        &\min_{w} \tfrac{C}{2}\|w\|^2_2 +\!\tfrac{1}{n}\!\sum_i
        \max_{y\in\mathcal{Y}} L(y,y_i)\left(1+f(y)-f(y_i)\right) 
\end{align}}

Most of the existing training algorithms for structural SVM require solving the maximization-over-labellings problems in \eqref{unconstrained_obj_slack} and \eqref{unconstrained_obj_margin}:
\begin{align}
\label{eq:margin_argmax}
\begin{array}{c}
\text{Margin}\\
\text{rescaling}
\end{array}:&~\argmax_{y\in\mathcal{Y}} L(y,y_i)+f(y)-f(y_i) \\
\label{eq:slack_argmax}
\begin{array}{c}
\text{Slack}\\
\text{rescaling}
\end{array}:&~\argmax_{y\in\mathcal{Y}} L(y,y_i)\left (1+f(y)-f(y_i)\right )
\end{align}

To better understand the difference between margin and slack rescaling
we focus on a single training instance $i$ and define the functions:
$h(y)=1+f(y)-f(y_i)$ and $g(y)=L(y_i,y)$. With these definitions we
see that the maximization \eqref{eq:margin_argmax} for margin
rescaling is $\max_{y\in\mathcal{Y}} h(y)+g(y)$, while the
maximization \eqref{eq:slack_argmax} for slack rescaling is
$\max_{y\in\mathcal{Y}} h(y)g(y)$.  It is now obvious why margin
rescaling is often easier. When the score and error
functions $h$ and $g$ \emph{decompose} into a sum of simpler
functions, we can exploit that structure in order to
solve the maximization efficiently
\cite{tsochantaridis2004support,Taskar03,finley2008training}.  In
contrast, the slack rescaling score does not decompose even when both
$h$ and $g$ do.  What we show, then, is how to solve problems of the
form $\max_y h(y)g(y)$, and thus the maximization
\eqref{eq:slack_argmax}, having access only to an oracle for additive
problems of the form $\max_y h(y)+\lambda g(y)$. 

That is, we assume that we have access to a procedure, referred to as the \emph{$\lambda$-oracle}, which can efficiently solve the problem:
\begin{equation}\label{eq:lambda_oracle}
y_\lambda=\mathcal{O}(\lambda)
=\argmax_{y\in\mathcal{Y}}\mathcal{L}_\lambda(y)
\end{equation}
where $\mathcal{L}_\lambda(y)=h(y)+\lambda g(y)$.  This problem is
just a rescaling of \eqref{eq:margin_argmax}.  E.g.,~for linear
responses it is obtained by scaling the weight vector by $1/\lambda$.
If we can handle margin rescaling efficiently we can most likely
implement the $\lambda$-oracle efficiently.  This is also the oracle
used by \citet{sarawagi2008accurate}.
In Section \ref{sec:alg}, we show how to obtain a solution to the
slack-rescaling problem \eqref{eq:slack_argmax} using such a
$\lambda$-oracle.  Our method can be used as a subroutine in a variety
of training algorithms, and we demonstrate that it is more scalable
than previous methods.  However, we also show that this approach is
limited, since no procedure that only has access to a $\lambda$-oracle
can guarantee the quality of its solution, no matter how much time it
is allowed to run.

Therefore, we propose an alternative procedure that can access a more
powerful oracle, which we call the \emph{constrained $\lambda$-oracle}:
\begin{align}\label{eq:constrained_oracle}
  y_{\lambda,\alpha,\beta}=\mathcal{O}_{c}(\lambda,\alpha,\beta)=\max_{y\in\mathcal{Y},\; \alpha h(y)> g(y),\; \beta h(y)\le g(y)} \mathcal{L}_\lambda(y),  
\end{align}
where $\alpha,\beta \in\reals$.
This oracle is similar to the $\lambda$-oracle, but can additionally handle linear constraints on the values $h(y)$ and $g(y)$. In the sequel we show that in many interesting cases this oracle is not more computationally expensive than the basic one.
For example, when the $\lambda$-oracle is implemented as a linear program (LP), the additional constraints are simply added to the LP formulation and do not complicate the problem significantly.
Before presenting our algorithms for optimizing \eqref{eq:slack_argmax}, we first review the training framework in the next section.

\section{Optimization for Slack Rescaling}
In this section we briefly survey cutting plane and stochastic
gradient descent optimization for the slack rescaled objective
\eqref{obj_slack} and \eqref{unconstrained_obj_slack}.  This will be
helpful in understanding the difference between our approach and that
of prior work on the slack rescaled objective by
\citet{sarawagi2008accurate}.

The cutting plane algorithm was proposed for solving the structural
SVM formulation in
\cite{tsochantaridis2004support,joachims2009cutting}.  This algorithm
has also been used in previous work on slack rescaling optimization
\cite{sarawagi2008accurate,bauer2014efficient}.  The difficulty in
optimizing \eqref{obj_slack} stems from the number of constraints,
which is equal to the size of the output space $\mathcal{Y}$ (for each
training instance).  The cutting plane method 
maintains a small set of constraints and solves the optimization only
over that set.  At each iteration the active set of constraints is
augmented with new violated constraints, and it can be shown that 
not too many such
constraints need to be added for a good solution to be found
\cite{joachims2009cutting}.  The main computational bottleneck here is to find
a violating constraint at each iteration, which is challenging since it
requires searching over the output space for some violating labeling
$y$. 

Relying on this framework, \citet{sarawagi2008accurate} use the formulation in \eqref{obj_slack} and rewrite the constraints as:
\begin{equation}
1 + f(y) - f(y_i) - \dfrac{\xi_i}{L(y,y_i)} \leq 0 \quad\forall i,y\ne y_i
\label{eq:scaled_constraint_1}
\end{equation}
Hence, to find a violated constraint they attempt to maximize: solve the problem with substitution of:
\begin{align}
\argmax_{y\in \mathcal{Y}'} \left ( h(y)-\dfrac{\xi_i}{g(y)}\right)
\label{yhat_c}
\end{align}
where $\mathcal{Y}'=\{y|y\in
\mathcal{Y},h(y)>0, 
y \neq y_i\}$ and we use our notation $h(y)=1 + f(y) - f(y_i)$ and $g(y)=L(y,y_i)$.
They suggest minimizing a convex upper bound of \eqref{yhat_c} which stems from the convex conjugate function of $\dfrac{\xi_i}{g(y)}$:
\begin{align}
&\max_{y\in \mathcal{Y}'} h(y)-\dfrac{\xi_i}{g(y)}
=\max_{y\in \mathcal{Y}'} \min_{\lambda\ge 0} \left ( h(y)+\lambda g(y)-2\sqrt{\xi_i \lambda}\right) \nonumber\\
\le &\min_{\lambda\ge 0} \underset{y\in \mathcal{Y}'}{\max} F'(\lambda,y) 
=\min_{\lambda\ge 0} \underset{y\in \mathcal{Y}'}{\max}\; F'(\lambda,y) = \min_{\lambda\ge 0} F(\lambda) \label{eq:p1_obj_up}
\end{align}
 where $F(\lambda)=\underset{y\in \mathcal{Y}'}{\max}\; F'(\lambda,y) =\max_{y\in \mathcal{Y}'} h(y)+\lambda g(y)-2\sqrt{\xi_i \lambda}$.
Since $F(\lambda)$ is a convex function, \eqref{eq:p1_obj_up} can be solved by a simple search method such as golden search over $\lambda$ \cite{sarawagi2008accurate}.

Although this approach is suitable for the cutting plane algorithm,
unfortunately it cannot be easily extended to other training
algorithms.  In particular, $F'(\lambda,y)$ is defined in terms of
$\xi_i$, which ties it to the constrained form \eqref{obj_slack}.  On
the other hand, algorithms such as stochastic gradient descent (SGD)
\cite{RatliffSubgradient,shalev2011pegasos}, stochastic dual
coordinate ascent (SDCA) \cite{shalev2013accelerated}, or
block-coordinate Frank-Wolfe (FW) \cite{lacoste2013block}, all
optimize the unconstrained objective form
\eqref{unconstrained_obj_slack}.
These methods are typically preferable in the large scale setting, since they have very low per-iteration cost, handling a single example at a time, with the same overall iteration complexity as cutting plane methods.
In contrast, the cutting plane algorithm considers the entire training set at each iteration, so the method does not scale well to large problems.
Since our goal in this work is to handle large datasets, we would like
to be able to use the stochastic methods mentioned above, working on
the  unconstrained formulation \eqref{unconstrained_obj_slack}.
The update in these algorithms requires solving the maximization
problem \eqref{eq:slack_argmax}, which is the goal of
\secref{sec:alg}.  Note that solving \eqref{eq:slack_argmax}
also allows using a cutting plane method if desired.

\section{Algorithms}
\label{sec:alg}

In this section we present our main contribution, a framework for
solving the maximization problem \eqref{eq:slack_argmax}, which we
write as:
\begin{align}\label{eq:Phi}
\max_y \Phi(y) := \max_y h(y) g(y)
\end{align}
We describe two new algorithms to solve this problem using access to the $\lambda$-oracle, which have several advantages over previous approaches.
However, we also show that any algorithm which uses only the $\lambda$-oracle cannot always recover an optimal solution.
Therefore, in \secref{sec:angular_w_constrained_oracle} we proposed an improved algorithm which requires access to an augmented $\lambda$-oracle that can also handle linear constraints.

\subsection{Binary search}
\label{sec:binary_search}
We first present a binary search algorithm similar to the one proposed by \citet{sarawagi2008accurate}, but with one main difference.
Our algorithm can be easily used with training methods that optimize the unconstrained objective \eqref{unconstrained_obj_slack}, and can therefore be used for SGD, SDCA and FW.
The algorithm minimizes a convex upper bound on $\Phi$ without slack variable $\xi_i$.
The algorithm is based on the following lemma (details and proofs are in \ref{app:quadratic_bound}).
\begin{lemma}\label{binary_upper}
Let $\bar{F}(\lambda)= \frac{1}{4}  \max_{y\in\mathcal{Y}^+}\left ( \frac{1}{\lambda}h(y)+\lambda g(y)  \right )^2$, then
\begin{align*} 
\max_{y\in\mathcal{Y}} \Phi(y)&
\le \min_{\lambda>0}\bar{F}(\lambda)
\end{align*}
and $\bar{F}(\lambda)$ is a convex function in $\lambda$. 
\end{lemma}

Rather than minimizing this upper bound, we next present an algorithm that aims to optimize $\Phi(y)$ in a more direct manner, using a geometrical interpretation of mapping labels into $\reals^2$.

\subsection{Geometrical Interpretation of $\lambda$-oracle search}

To understand the problem better and motivate our methods, it is
useful to consider the following geometrical interpretation of
\eqref{eq:Phi}: we map each labels $y$ to a vector $\vec{y}=[h(y) \;
g(y)] \in \reals^2$.  Let $\vec{\mathcal{Y}}=\{\vec{y}\in \reals^2|y\in
\mathcal{Y}\}$ be the set of the all mapped labels.  The maximization
\eqref{eq:Phi} reduces to the problem: given a set of points
$\vec{\mathcal{Y}}\subset \reals^2$, maximize the product of their coordinates
$\vec{y^*}=\argmax_{\vec{y}\in\vec{\mathcal{Y}}}[\vec{y}]_1\cdot[\vec{y}]_2$.

\begin{figure}
\begin{subfigure}[b]{.49\linewidth}
        \includegraphics[width=\linewidth]{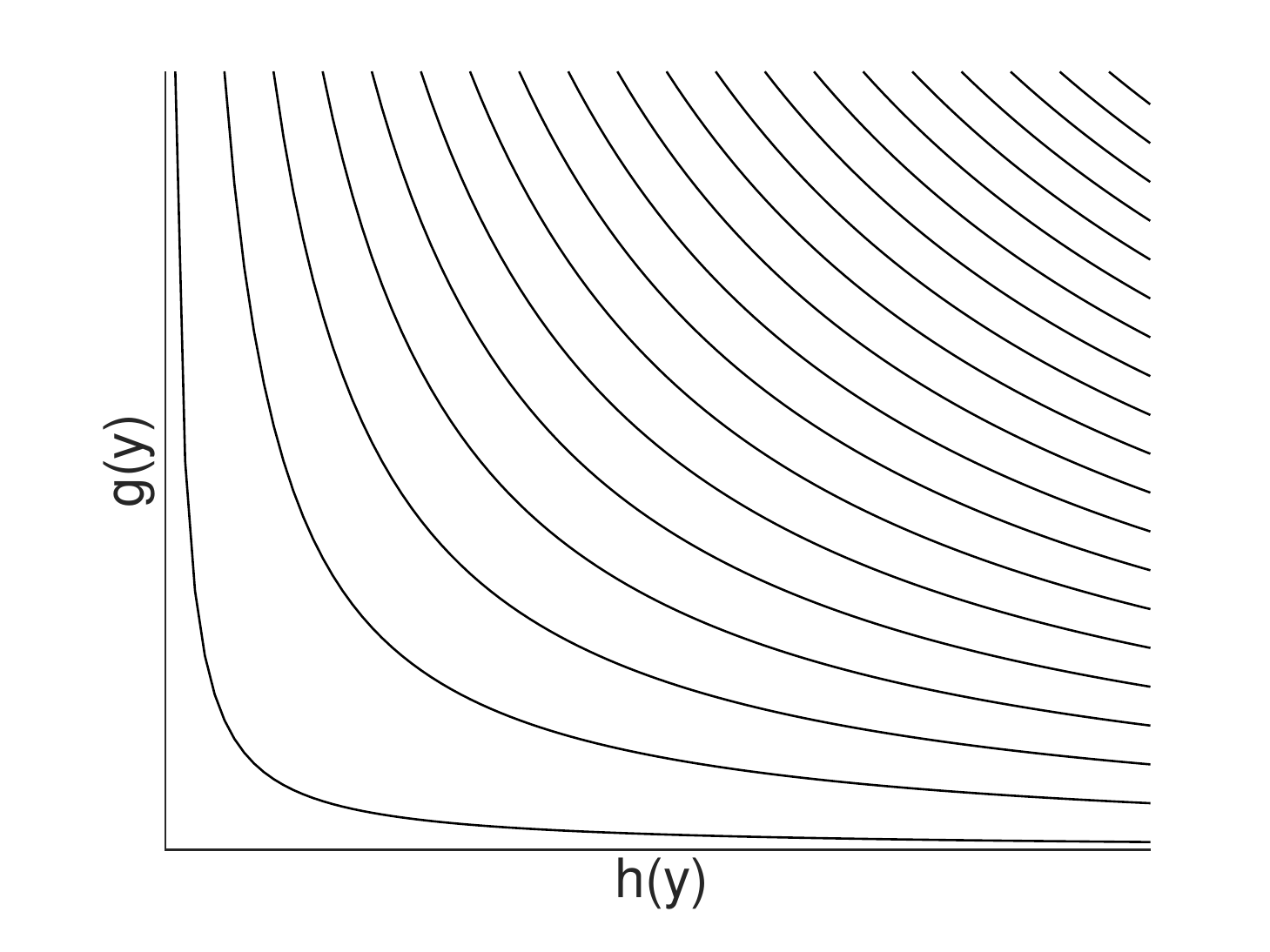}
        \label{fig:phi_contour}
        \caption{$\Phi$ contour.}
    \end{subfigure}
\begin{subfigure}[b]{.49\linewidth}
        \includegraphics[width=\linewidth]{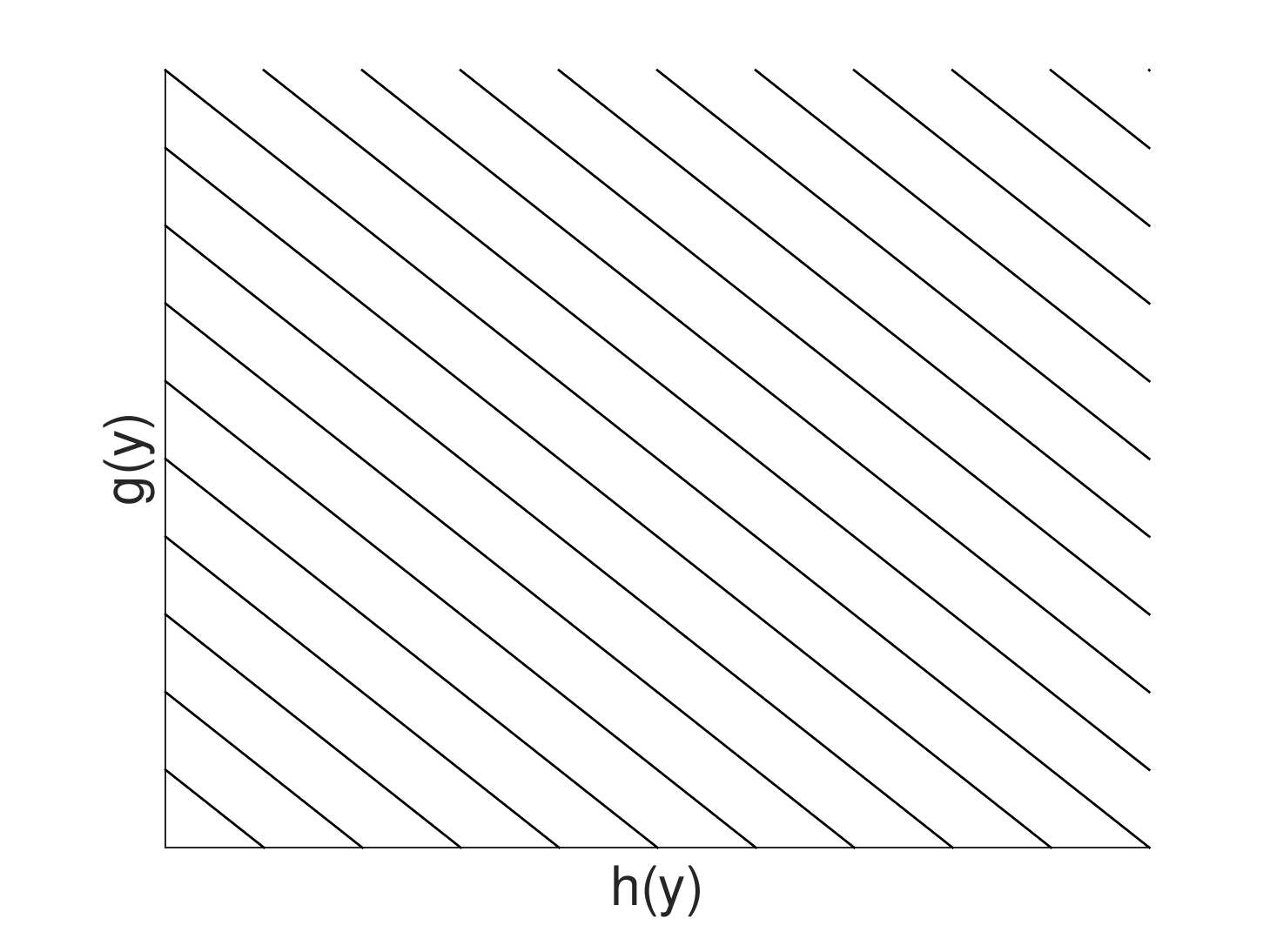}
        \label{fig:lambda_contour}
        \caption{$\lambda$ contour.} 
    \end{subfigure}    
\caption{Contour of the two functions considered in $\reals^2$. $\Phi$ contour is the contour of the objective function, and $\lambda$ contour is the contour used by the oracle. }\label{fig:contour}
\end{figure}

The contours of our objective function
$\vec{\Phi}(\vec{y})=[\vec{y}]_1\cdot [\vec{y}]_2$ are then
hyperbolas.  We would like to maximize this function by repeatedly
finding points that maximize linear objectives of the form
$\vec{\mathcal{L}}_\lambda(\vec{y})=[\vec{y}]_1+\lambda [\vec{y}]_2$,
whose contours form lines in the plane.  See Figure \ref{fig:contour}.

An example of mapping of label into $\reals^2$ is shown in \ref{ap:map}.

\begin{figure}
\centering     
\includegraphics[width=\linewidth]
{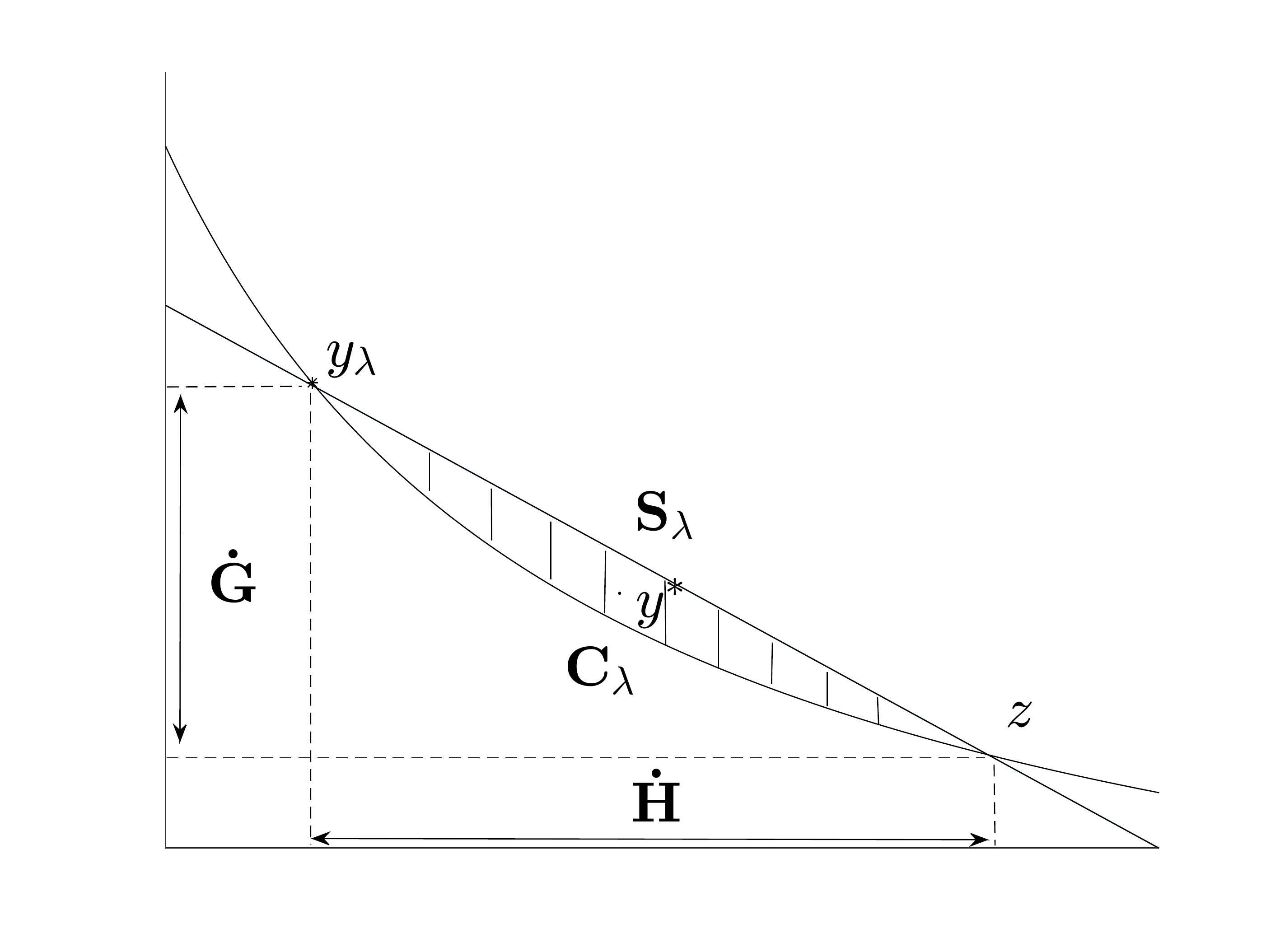}
\caption{Geometric interpretation of the $\lambda$-oracle: $\vec{y^*}$ must reside between the upper bound $S_\lambda$ and the lower bound $C_\lambda$, the shaded area. It follows that $h(y^*)$ and $g(y^*)$ reside in a simple segment  $\dot{H}$ and $\dot{G}$ respectively.}\label{fig:geo}
\end{figure}
The importance of the $\reals^2$ mapping is that each $y_\lambda$
revealed by the $\lambda$-oracle shows that $y^*$ can only reside in a
small slice of the plane.
See figure \ref{fig:geo}.

\begin{lemma}\label{lem:restricted_space}
Let $S_\lambda$ be a line through $\vec{y}_\lambda$ and $\vec{z}=[\lambda[\vec{y}_\lambda]_2,\frac{1}{\lambda}[\vec{y}_\lambda]_1]$, and   let $C_{\lambda}=\{\vec{y}\in\reals^2|[\vec{y}]_{1}\cdot [\vec{y}]_{2}=\vec{\Phi}(\vec{y}_{\lambda})\}$ be the hyperbola through $\vec{y}_\lambda$.
Then, 
$\vec{y^*}$  is on or below line $S_\lambda$, and $\vec{y^*}$  is on or above hyperbola $C_\lambda$. 
\proof 
If there exists a $\vec{y}\in\vec{\mathcal{Y}}$ which is above $S_\lambda$, it contradicts the fact that $\vec{y}_\lambda$ is the argmax point for function $\vec{\mathcal{L}}_\lambda$.  And the second argument follows from $\vec{y^*}$ being the argmax label w.r.t.~$\vec{\Phi}$, and the area above $C_\lambda$ corresponds to points whose $\vec{\Phi}$ value is greater than $\vec{y_\lambda}$. \qed
\end{lemma}
 It follows that $h(y^*)$ and $g(y^*)$ must each reside in a segment:
\begin{lemma} \label{cor:simple_segment}
Let $\dot{H}=[ \min([\vec{y_\lambda}]_1,[\vec{z}]_1), \max([\vec{y_\lambda}]_1,[\vec{z}]_1)]$ and $\dot{G}=[\min([\vec{y_\lambda}]_2,[\vec{z}]_2), \max([\vec{y_\lambda}]_2,[\vec{z}]_2)]$.
Then, 
\begin{align*}
h(y^*)\in \dot{H}, &&g(y^*)\in \dot{G}
\end{align*} 
\proof This follows from the fact that $S_\lambda$ and $C_\lambda$ intersects at two points, $\vec{y_\lambda}$ and $\vec{z}$, and the boundaries, $S_\lambda$ and $C_\lambda$, are strictly decreasing functions. \qed
\end{lemma}

\subsection{Bisecting search}
\label{sec:bisecting_search}

In this section, we propose a search algorithm which is based on the previous geometric interpretation.
Similar to the binary search, our method also relies on the basic $\lambda$-oracle. 

We next give an overview of the algorithm. We maintain a set of  possible value ranges $\lambda^*=\argmax_{\lambda>0}\Phi(y_\lambda)$, $h(\lambda^*)$, and $g(\lambda^*)$ as $L, H,$ and $G$, respectively; all initialized as  $\reals$.
First, for each $y_\lambda$ returned by the oracle,  we take an intersection of $G$ and $H$ with a segment of possible values of $h(y)$ and $g(y)$, respectively, using Lemmas \ref{lem:restricted_space} and \ref{cor:simple_segment}.
%
Second, we reduce the space $L$ of potential $\lambda$'s based on the following Lemma (proved in the Appendix).
\begin{lemma}\label{MAP_seg}
$h(y_\lambda)$ is a non-increasing function of $\lambda$, and $g(y_\lambda)$
is a non-decreasing function of $\lambda$.
\end{lemma}
Thus, we can discard $\{\lambda'|\lambda'>\lambda\}$ if $h(y^*_\lambda)>h(y_\lambda)$ or $\{\lambda'|\lambda'<\lambda\}$ otherwise from $L$.
Next, we pick $\lambda\in L$ in the middle, and query $y_\lambda$.
The algorithm continues until at least one of $L, H,$ and $G$ is empty.

Similar to the binary search from the previous section, this algorithm can be used with training methods like SGD and SDCA, as well as the cutting-plane algorithm. 
However, this approach has several advantages compared to the binary search.
First, the binary search needs explicit upper and lower bounds on $\lambda$, thus it has to search the entire $\lambda$ space \cite{sarawagi2008accurate}.
However, the bisecting search can directly start from any $\lambda$ without an initial range, and for instance, this can be used to warm-start from the optimal $\lambda$ in the previous iteration. 
Furthermore, we point out that since the search space of $h$ and $g$ is also bisected, the procedure can terminate early if either of them becomes empty. 
 
Finally, in \ref{app:search_improvements} we propose two improvements that can be applied to either the binary search or the bisecting search. Specifically, we first provide a simple stopping criterion that can be used to terminate the search when the current solution $y_{\lambda_t}$ will not further improve. Second, we show how to obtain a bound on the suboptimality of the current solution, which can give some guarantee on its quality.

So far we have used the $\lambda$-oracle as a basic subroutine in our search algorithms.
Unfortunately, as we show next, this approach is limited as we cannot guarantee finding the optimal solution $y^*$, even with unlimited number of calls to the $\lambda$-oracle.
This is somewhat distressing since with unlimited computation we can find the optimum of \eqref{eq:slack_argmax} by enumerating all $y$'s.

\begin{algorithm}
  \caption{Bisecting search }
\label{alg:bisecting}
  \begin{algorithmic}[1]
    \Procedure{Bisecting}{$\lambda_0$}
    \INPUT {Initial $\lambda$ for the search $\lambda_0\in\mathbb{R}_+$
}
    \OUTPUT{$\hat{y} \in \mathcal{Y}.$}
    \INIT  {$H=G=L=\mathbb{R}_+,\lambda=\lambda_0$, $\hat{\Phi}=0$.}
    \While {$H\ne\emptyset$ and $G\ne\emptyset$ }
     \State $y'\gets \mathcal{O}(\lambda)$
     \State $u\gets [h(y')\; \lambda g(y')],v \gets [g(y')\; \frac{1}{\lambda} h(y')]$
     \State $H\gets H \cap \{h'|\min u\le h' \le \max u\}$ \Comment{Update}
     \State $G\gets G \cap \{g'|\min v\le g' \le \max v\}$
     \If{$v_{1}\le v_2$}\Comment{Increase $\lambda$}
       \State $L\gets L \cap \{\lambda'\in\mathbb{R}|\lambda'\ge\lambda\}$
     \Else\Comment{Decrease $\lambda$}
       \State $L\gets L \cap \{\lambda'\in\mathbb{R}|\lambda'\le \lambda\}$
     \EndIf
    \State $\lambda \gets \frac{1}{2}(\min L+\max L)$
    \If {$h(y')g(y')\ge\hat{\Phi}$}
     \State $\hat{y}\gets y',$ $\hat{\Phi}\gets h(y')g(y')$.
    \EndIf
    \EndWhile
    \EndProcedure
  \end{algorithmic}
\end{algorithm}


\subsection{Limitation of the $\lambda$-oracle}
\label{sec:lambda_oracle_inexact}
Until now, we used only the $\lambda$-oracle to search for $\Phi^*$ without
directly accessing the functions $h$ and $g$. We now show that this
approach, searching $\Phi^*$ with only a $\lambda$-oracle, is very limited:
even with an unlimited number of queries, the search cannot be
exact and might return a trivial solution in the worst case (see
\ref{lambda_oracle_proof} for proof).
\begin{theorem}\label{loracle_l1}
  Let $\hat{H}=\max_y h(y)$ and $\hat{G}=\max_y g(y)$. For any
  $\epsilon>0$, there exists a problem with 3 labels
  such that for any $\lambda\geq0$, $\Phi(y_\lambda)=\min_{y\in \mathcal{Y}} \Phi(y)<\epsilon$, while
  $\Phi(y^*)=
  \dfrac{1}{4}\hat{H}\hat{G}$. 

 \end{theorem}
 Theorem \ref{loracle_l1} shows that any search algorithm that can access
 the function only through $\lambda$-oracle, including the method of
 \citet{sarawagi2008accurate} and both methods presented above,
 cannot be guaranteed to find a label optimizing $\Phi(y)$, even
 approximately, and even with unlimited accesses to the oracle. This
 problem calls for a more powerful oracle.

\subsection{Angular search with the constrained-$\lambda$-oracle}
\label{sec:angular_w_constrained_oracle}


The \emph{constrained $\lambda$-oracle}   defined in
\eqref{eq:constrained_oracle} has two inequality constraints to
restrict the search space.  Using this modified algorithm, we can
present an algorithm that {\em is} guranteed to find the most
violating constraint, as captured by the following guarantee, proved
in \ref{angular_search_proof}:
\begin{theorem}\label{thm:angular_optimal} Angular search described in
  Algorithm \ref{alg:angular} finds the optimum $y^*=\argmax_{y\in
    \mathcal{Y}} \Phi(y) $ using at most $t=2M+1$ iteration where $M$ is the number of the labels.
\end{theorem}
This is already an improvement over the previous methods, as at least
we are guaranteed to return the actual most violating label.  However,
it is still disappointing since the number of iterations, and thus
number of oracle accesses might actually be larger than the number of
labels.  This defies the whole point, since we might as well just
enumerate over all $M$ possible labels.  Unfortunately, even with a
constrained oracle, this is almost the best we can hope for. In fact,
even if we allow additional linear constraints, we might still need
$M$ oracle accesses, as indicated by the following Theorem, proved in \ref{lambda_oracle_proof}.

\begin{theorem}\label{loracle_l2}
Any search algorithm accessing labels only through a $\lambda$-oracle with any number of linear constraints cannot find $y^*$ using less than $M$ iterations in the worst case, where $M$ is the number of labellings.
\end{theorem}

Fortunately, even though we cannot guarantee optimizing $\Phi(y)$
exactly using a small number of oracle accesses, we can at least do so
approximately. This can be achieved by Algorithm \ref{alg:angular} (see \ref{angular_search_proof}), as the next theorem states.

\begin{theorem}\label{thm:angular_suboptimality} In angular search, described in Algorithm \ref{alg:angular}, at iteration $t$, 
\begin{align*}
\dfrac{\Phi(y^*)}{\Phi(\hat{y}^t)}\le (v_1)^{\frac{4}{t+1}}
\end{align*}
where $\hat{y}^t = \argmax_t y^t$ is the optimum up to $t$, $v_1=\max \left \{\dfrac{\lambda_0}{\partial(y_1)}, \dfrac{\partial(y_1)}{\lambda_0}\right\}$, $\lambda_0$ is the initial $\lambda$ used, and $y_1$ is the first label returned by constrained $\lambda$-oracle.
\end{theorem}
We use $\partial(a)=\frac{a_2}{a_1}$ to denote the slope of a vector.

With proper initialization, we get the following runtime guarantee:
\begin{theorem}\label{thm:angular_running_time}  Assuming $\Phi(y^*)>\phi$, angular search described in algorithm \ref{alg:angular} with $\lambda_0=\dfrac{\hat{G}}{\hat{H}}, \alpha_0=\dfrac{\hat{G}^2}{\phi},\beta_0=\dfrac{\phi}{\hat{H}^2}$, finds an $\epsilon$-optimal solution, $\Phi(y)\ge(1-\epsilon)\Phi(y^*)$, in $T$ queries and $O(T)$ operations, where $T=4\log\left(\dfrac{\hat{G}\hat{H}}{\phi}\right)\cdot \dfrac{1}{\epsilon}$, and $\delta$-optimal solution, $\Phi(y)\ge\Phi(y^*)-\delta$,  in $T'$ queries and $O(T')$ operations, where $T'= 4\log\left(\dfrac{\hat{G}\hat{H}}{\phi}\right)\cdot \dfrac{\Phi(y^*)}{\delta}$.
\end{theorem}

 \begin{figure}
\includegraphics[trim = 50mm 0mm 50mm 0mm, clip, 
width=70mm]
{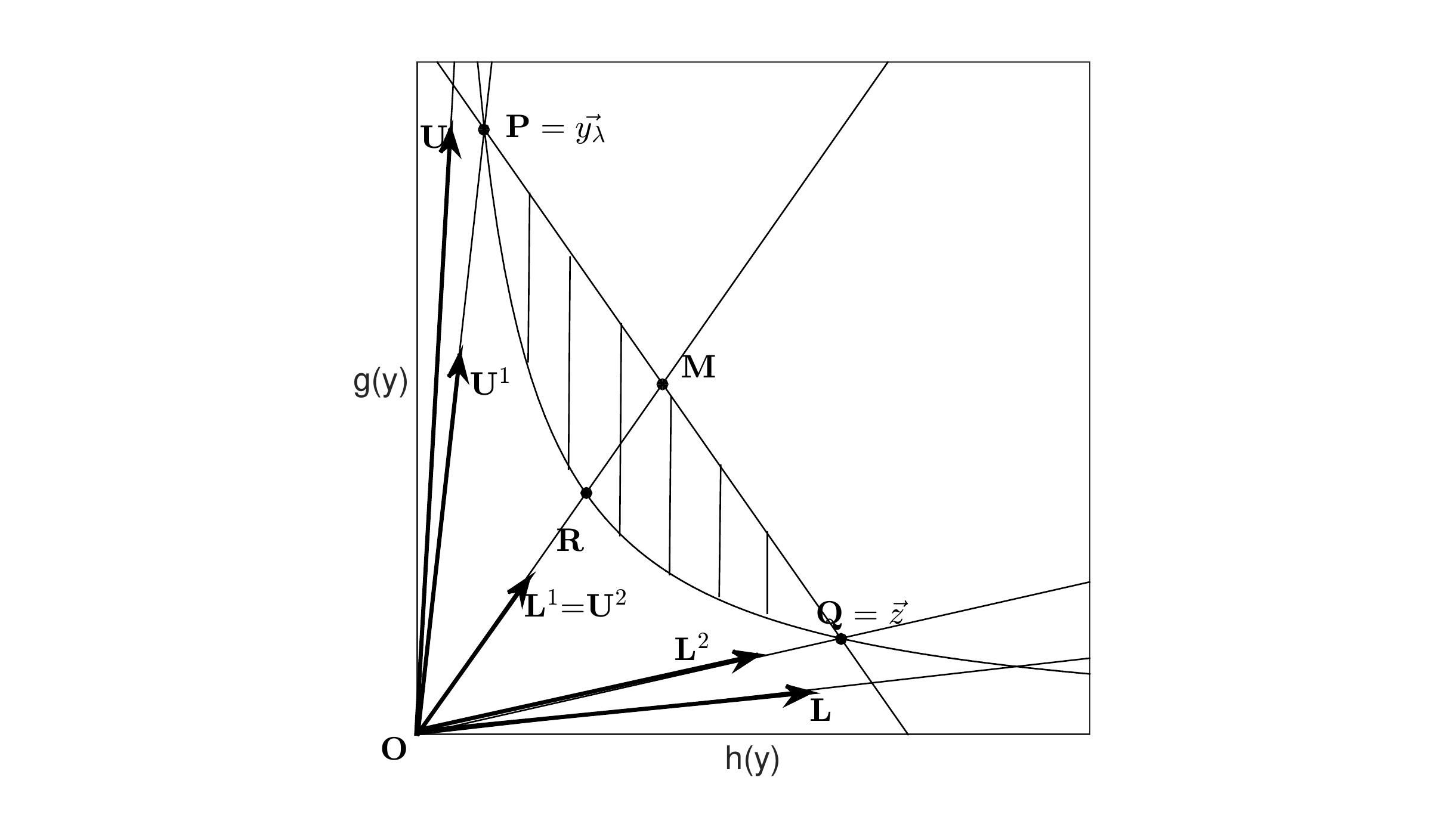}
\caption{Split procedure.}
\label{fig:split_cor}
\end{figure}

Here we give an overview of the algorithm with an illustration in Figure \ref{fig:split_cor}. The constrained $\lambda$-oracle restricts the search space, and this restriction can be illustrated as a linear upper bound $U$ and a lower bound $L$.
The search is initialized with the entire right angle: $U=[0\; \infty]$ and $L=[\infty\; 0]$, and maintains that $\vec{y^*}$ is always between $U$ and $L$. 
The constrained $\lambda$-oracle is used with $U,L$ and a certain $\lambda$ to reduce the potential area where $\vec{y^*}$ can reside.
Specifically, the search space is reduced using an angle defined by $U=\overline{OP}$ and $L=\overline{OQ}$.
In the next iteration, the constrained $\lambda$-oracle is invoked with $U^1=\overline{OP}$ and $L^1=\overline{OM}$, and also with $U^2=\overline{OM}$ and $L^2=\overline{OQ}$.
Intuitively, each such query shrinks the search space, and as the search space shrinks, the suboptimaly bound improves.
This process is continued until the remaining search space is empty.
The angular search algorithm defines the optimal $\lambda$ and values to be passed to the constrained $\lambda$-oracle.

In Algorithm \ref{alg:angular} each angle is dequeued, split, and enqueued recursively.
Each angle maintains its upper bound from the previous iterations and stops splitting itself and terminate if it is ensured that there exists no label with larger $\Phi$ value within the angle. 
When the oracle reveals a label with $\Phi(y_\lambda)=c$, we can safely discard all area corresponding to $\{\vec{y}|\Phi(\vec{y})\le c\}$. 
This works as a global constraint which shrinks the search space. Therefore, acquiring a label with high $\Phi$ value in the early stages facilitate convergence. Thus, it is suggested to use a priority queue, and dequeue the angle with the highest upper bound on $\Phi$. 
A similar strategy is to have a {\em label cache}, the subset of previous most violated labels, denoted as $\mathcal{C}$.  With the label cache, we can discard a large part of the search space $\{\vec{y}|\Phi(\vec{y})\le \max_{y'\in \mathcal{C}}\Phi(\vec{y'})\}$ immediately.
Algorithm \ref{alg:angular} also uses the constrained $\lambda$-oracle to avoid returning previously found labels.
Finally, for $\lambda_0$, we suggest to use $\lambda_0=\frac{\hat{H}}{\hat{G}}$, with $\hat{H}$ calculated from the current weights $w$. 


 
%
 See \ref{ap:illu} for the illustration of the angular search.
\section{Experiments}
\label{sec:experiments}
 In this section, we validate our contributions by comparing the different behaviors of the search algorithms on standard benchmark datasets, and its effect on the optimization. Specifically, we show that angular search with SGD is not only much faster than the other alternatives, but also enables much more precise optimization with slack rescaling formulation, outperforming margin rescaling. 

Unlike the simple structure used in \cite{bauer2014efficient}, we show applicability to complicated structure. We experiment with multi-label dataset modeled by a Markov Random Field with pair-wise potentials as in \cite{finley2008training}. Since the inference of margin scaling is NP-hard in this case, we rely on linear programming relaxation. Note that this complicates the problem, and  the number of labels becomes even larger (adds fractional solutions). 
Also notice that all of our results in previous sections apply with mild modification to this harder setting.
Two standard benchmark multi-label datasets, yeast\cite{elisseeff2001kernel} (14 labels)
 and RCV1\cite{lewis2004rcv1}, are tested.
 For RCV1 we reduce the data to the $50$ most frequent labels.
 For angular search, we stop the search whenever $\frac{\Phi(\hat{y})}{\Phi(y^*)}>0.999$ holds, to avoid numerical issues.
 \subsection{Comparison of the search algorithms}
 
  \begin{table}[h!]
  \begin{center}
    \begin{tabular}{| l  | c |c|c|}
    \hline
      &  Angular & Bisecting & Sarawagi \\
      \hline            \hline
 \multicolumn{4}{|c|}{Yeast (N=160)} \\      
    \hline
    Success &  22.4\%\ & 16.5\%\ & 16.4\% \\
    Queries per search & 3.8 & 10.3 & 43.2 \\
  Average time (ms) & 4.7 & 3.6 & 18.5 \\
      \hline            \hline
 \multicolumn{4}{|c|}{RCV1 (N=160)} \\      
    \hline
    Success &   25.6\%\  &      18.2\%\    &     18\%\\
    Queries per search &    4.8    &         12.7 \   &          49 \\
 Average time (ms) & 4.4 & 5.2 & 20.9 \\
    \hline    
    
    \end{tabular}\\
  \end{center}
  
  \caption{Comparison of the search algorithm. }\label{tabel_argmax}
\end{table} 
 Table \ref{tabel_argmax} compares the performance of the search in terms of the time spend, the number of queries, and the success percentage of finding the most violating label. 
The cutting-plane algorithm calls the search algorithms to find the most violating label $\hat{y}$, and adds it the active set if the violation is larger than some margin $\epsilon$, i.e.,  $ \Delta(\hat{y},y_i)(1+f(\hat{y})-f(y_i))>\xi_i+\epsilon$.
For cutting-plain optimization, we compare all three algorithms: Angular search, Bisecting search, and Sarawagi and Gupta's \cite{sarawagi2008accurate} (but just used Angular search for the update).
Success percentage is the percentage that the search algorithm finds such a violating label.
As expected from Theorem \ref{loracle_l1}, bisecting and Sarawagi's search miss the violating label in cases where angular search successfully finds one.
This is important for obtaining high accuracy solution. 
For RCV1 dataset, not only is angular search more accurate, but it also uses about 2.6 times less queries than bisecting and 10.1 times less queries than Sarawagi's search.
As for the timing, angular search is 1.18 times faster than bisecting search, and 4.7 times faster than Sarawagi's algorithm. 
 
\begin{figure}
\centering     
\begin{subfigure}[b]{.7\linewidth}
\includegraphics[width=\linewidth]{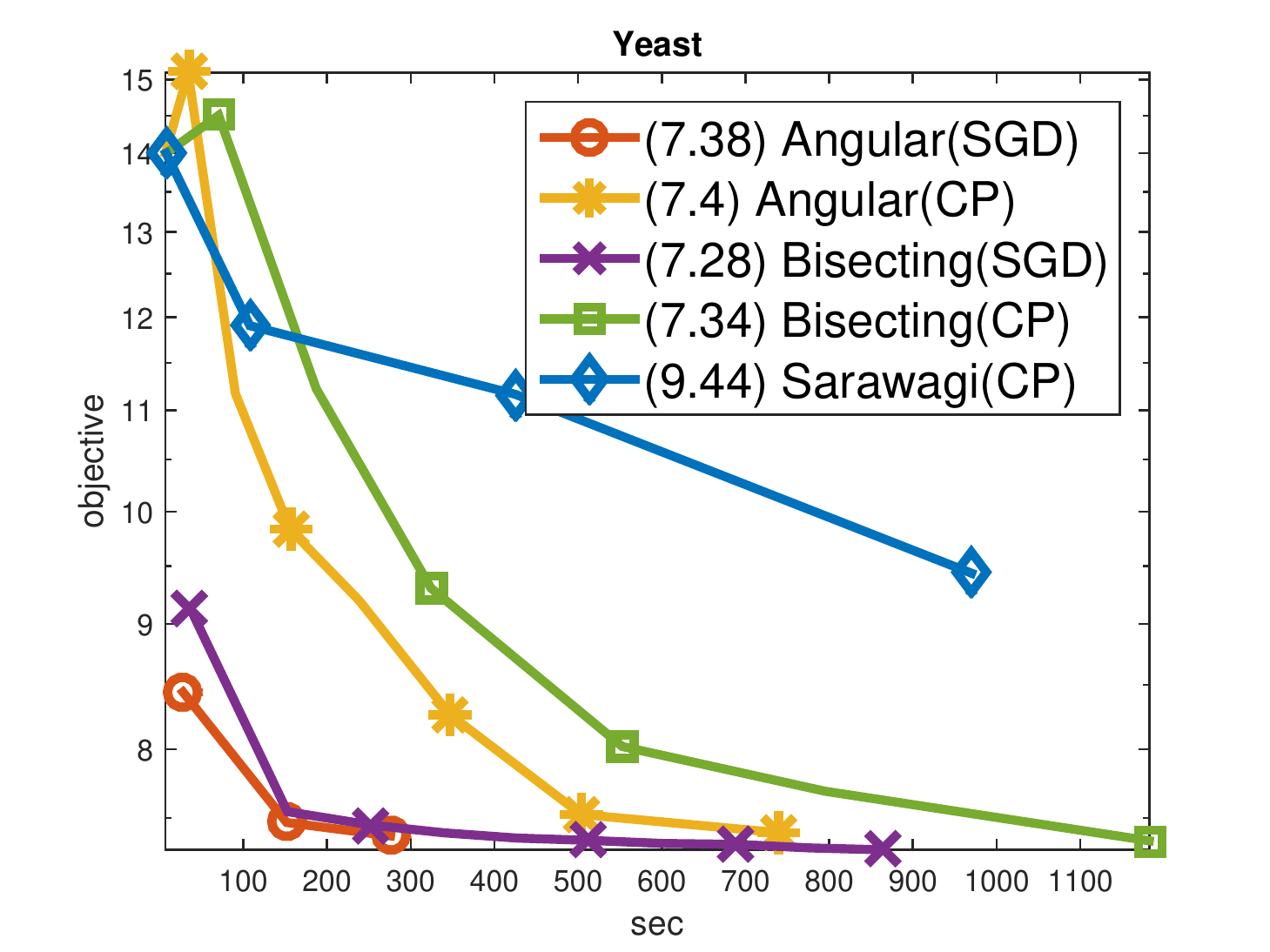}
\caption{Yeast Convergence rate}
\end{subfigure}
\begin{subfigure}[b]{.7\linewidth}
\includegraphics[width=\linewidth]{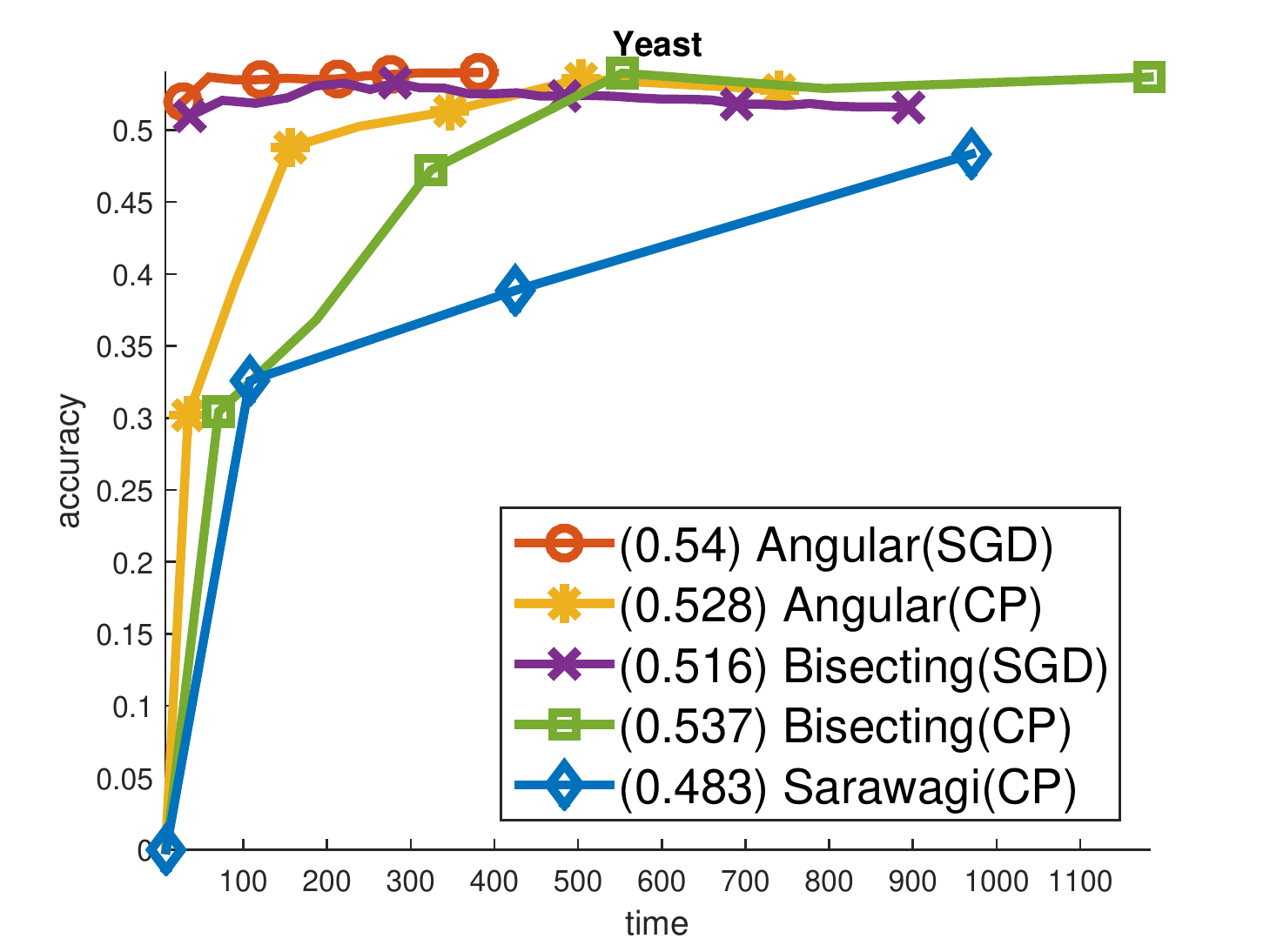}
\caption{Yeast Accuracy}
\end{subfigure}
\begin{subfigure}[b]{.7\linewidth}
\includegraphics[width=\linewidth]{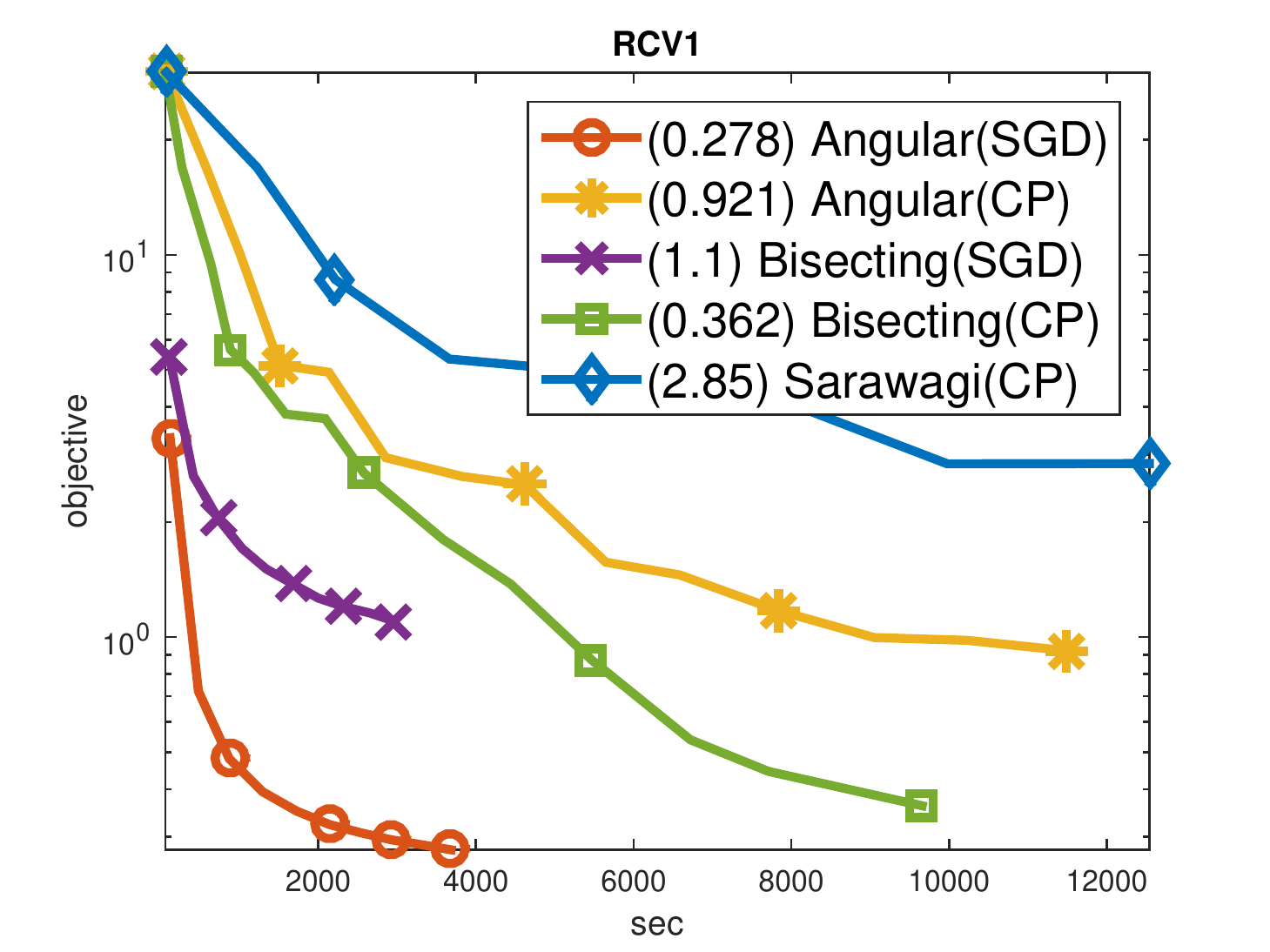}
\caption{RCV1 Convergence rate}
\end{subfigure}
\begin{subfigure}[b]{.7\linewidth}
\includegraphics[width=\linewidth]{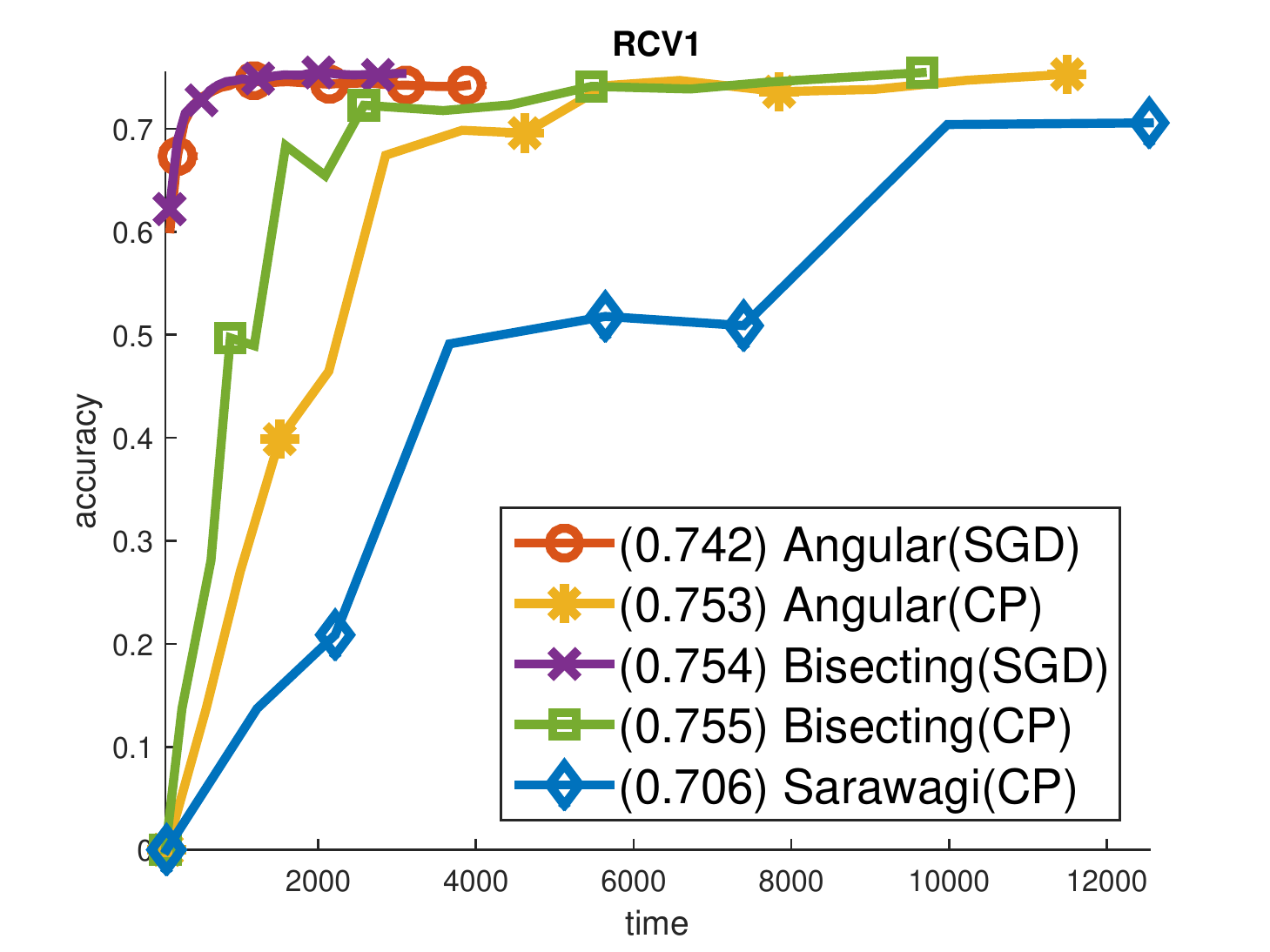}
\caption{RCV1 Accuracy}
\end{subfigure}
\caption{Convergence rate and the accuracy. Angular search with SGD is significantly faster and performs  the others. }
\label{fig:convergence}
\end{figure}

 In figure \ref{fig:convergence}, we compare the convergence rate and the accuracy of the different optimization schemes using different  search algorithms. Additional plots showing convergence w.r.t.~the number of queries and iterations are in \ref{ap:additional_plot}.  These show that angular search with SGD converges order of magnitude faster.
 
 Table \ref{table:result} shows a performance comparison for the multi-label datasets.
 For RCV1 dataset it shows a slight performance gain, which shows that the benefit of slack rescaling formulation is greater when the label space is large. 

%
%
\begin{table}
\centering
Yeast

    \begin{tabular}{| l  | c |c|c|c|}
    \hline
&  Acc & Label loss & MiF1 & MaF1\\
    \hline
    Slack  
      & .54  & .205 &  .661 &  .651\\
     Margin 
      &  .545& .204 &  .666 &  .654\\
     \hline
    \end{tabular}

    RCV1
    
        \begin{tabular}{| l  | c |c|c|c|}
    \hline
&  Acc & Label loss & MiF1 & MaF1\\
    \hline
    Slack  
      & .676  & .023 &  .755 &  .747\\
     Margin 
      &  .662 & .023 &  .753 &  .742\\
     \hline
    \end{tabular}
\caption{Results on Multi-label Dataset with Markov Random  Field.}\label{table:result}
\end{table}

\subsection{Hierarchical Multi-label Classification}

 We further experimented on problem of hierarchical multi-label classification \cite{cai2004hierarchical}. 
 In hierarchical multi-label classification, each label $y$ is a leaf node in a given graph, and each label $y$ shares ancestor nodes. It can be described as a graphical model where a potential of a multi-label $Y=\{y_1,\dots,y_k\}$ is the sum of all potentials of its ancestors, i.e., $\Phi(Y)=\sum_{n\in \bigcup_{n\in Y} Anc(n)} \Phi(n)$. We extracted 
                    1500 instances with dimensionality 17944 with a graph structure of 156 nodes with 123 label from 
           SWIKI-2011. SWIKI-2011 is a multi-label dataset of wikipedia pages from 
LSHTC competition\footnote{http://lshtc.iit.demokritos.gr/}. We used 750 instances as training set, 250 instances as holdout set, and 750 instances as test set. The Hamming distance is used as label loss. We show that slack rescaling in such large label structure is tractable and outperforms margin rescaling.

  \begin{table}[h!]

    \begin{tabular}{| l  | c |c|c|c|}
    \hline
&  Acc & Label loss & MiF1 & MaF1\\
    \hline
    Slack  
      & .3798 & .0105 &  .3917 &  .3880\\
     Margin 
      &  .3327& .0110 &  .3394 &  .3378\\
     \hline
    \end{tabular}
  \caption{Result on hierarchical multi-label dataset}
\end{table}

\section{Summary}

As we saw in our experiments, and has also been previously noted,
slack rescaling is often beneficial compared to margin rescaling in
terms of predictive performance.  However, the margin-rescaled argmax
\eqref{eq:margin_argmax} is often much easier computationally due to its
additive form.  Margin rescaling is thus much more frequently used in
practice.  Here, we show how an oracle for solving an argmax of the
form \eqref{eq:margin_argmax}, or perhaps a slightly modified form (the
constrained-$\lambda$ oracle), is sufficient for also obtaining exact
solutions to the slack-rescale argmax \eqref{eq:slack_argmax}.  This
allows us to train slack-rescaled SVMs using SGD, obtaining better
predictive performance than using margin rescaling.  Prior work in
this direction \cite{sarawagi2008accurate} was only approximate, and
more significantly, only enabled using cutting-plane methods, {\em
  not} SGD, and was thus not appropriate for large scale problems.
More recently, \cite{bauer2014efficient} proposed an efficient
dynamic programming approach for solving the slack-rescaled argmax
\eqref{eq:slack_argmax}, but their approach is only valid for sequence
problems\footnote{The approach can also be generalized to tree-structured
  problems.} and only when using hamming errors, not for more general
structured prediction problems.  Here, we provide a generic method
relying on a simple explicitly specified oracle that is guaranteed to
be exact and efficient even when the number of labels are infinite and allows using SGD and thus working with large scale
problems.

\bibliography{slack_rescaling}
\bibliographystyle{apalike}
\appendix

\clearpage
\gdef\thesection{Appendix \Alph{section}}
\section{Details of binary search}
\label{app:quadratic_bound}
\begin{replemma}{binary_upper}
Let $\bar{F}(\lambda)= \frac{1}{4}  \max_{y\in\mathcal{Y}^+}\left ( \frac{1}{\lambda}h(y)+\lambda g(y)  \right )^2$, then
\begin{align*} 
\max_{y\in\mathcal{Y}} \Phi(y)&
\le \min_{\lambda>0}\bar{F}(\lambda)
\end{align*}
and $\bar{F}(\lambda)$ is a convex function in $\lambda$. 

\proof
First, let $\mathcal{Y}^+=\{y|y\in \mathcal{Y},h(y)>0\}$, then $\max_{y\in\mathcal{Y}} \Phi(y) = \max_{y\in\mathcal{Y}^+} \Phi(y)$,
since any solution $y$ such that $h(y) < 0$ is dominated by $y_i$, which has zero loss.
Second, we prove the bound w.r.t. $y\in\mathcal{Y}^+$.
In the following proof we use a quadratic bound (for a similar bound see \cite{nguyen1995minimizing}).
\begin{align}
\max_{y\in\mathcal{Y}^+} \Phi(y)&
=\max_{y\in\mathcal{Y}^+} h(y)g(y)= \max_{y\in\mathcal{Y}^+} \frac{1}{4}\left (2\sqrt{h(y)g(y)} \right )^2\nonumber\\
&=\frac{1}{4}\left ( \max_{y\in\mathcal{Y}^+} \min_{\lambda>0} \left \{\frac{1}{\lambda}h(y)+\lambda g(y) \right \}\right )^2\nonumber
 \\ &\le \frac{1}{4} \left (\min_{\lambda>0}\ \max_{y\in\mathcal{Y}^+} \left \{\frac{1}{\lambda}h(y)+\lambda g(y) \right \} \right )^2 \label{upper2}
\end{align}
To see the convexity of $\bar{F}(\lambda)$, we differentiate twice to obtain:
\begin{align*}
\dfrac{\partial^2 \bar{F}(\lambda)}{\partial \lambda^2}=\frac{1}{4}  \max_{y\in\mathcal{Y}^+} 6\frac{1}{\lambda^4}h(y)^2+2g(y)^2>0
\end{align*}
 \qed
\end{replemma} 
Similar to \cite{sarawagi2008accurate}, we obtain a convex upper bound
on our objective.
Evaluation of the upper bound $\bar{F}(\lambda)$ requires using only the $\lambda$-oracle.
Importantly, this alternative bound $\bar{F}(\lambda)$ does not depend on the slack variable $\xi_i$, so it can be used with algorithms that optimize the unconstrained formulation \eqref{unconstrained_obj_slack}, such as SGD, SDCA and FW.
As in \cite{sarawagi2008accurate}, we minimize $\bar{F}(\lambda)$ using \emph{binary search} over $\lambda$.
The algorithm keeps track of $y_{\lambda_t}$, the label returned by the $\lambda$-oracle for intermediate values $\lambda_t$ encountered during the binary search, and returns the maximum label $\max_t \Phi(y_{\lambda_t})$.
This algorithm focuses on the upper bound $\min_{\lambda>0}\bar{F}(\lambda)$, and interacts with the target function $\Phi$ only through evaluations $\Phi(y_{\lambda_t})$ (similar to \cite{sarawagi2008accurate}).

%
\section{An example of label mapping}\label{ap:map}
\begin{figure}[H]
\centering     
\includegraphics[width=\linewidth]{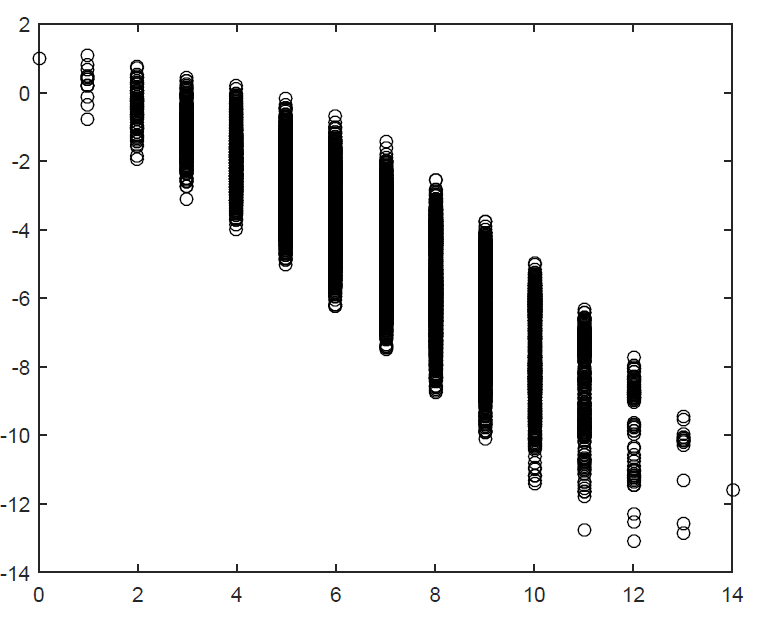}
\caption{A snapshot of labels during optimiation with Yeast dataset.  Each $2^{14}-1$ labels is shown as a point in the figure \ref{fig:all_points}. X-axis is the $\triangle(y,y_i)$ and Y-axis is $1+f_W(y)-f_W(y_i)$.}\label{fig:all_points}
\end{figure}
\section{Monotonicity of $h$ and $g$ in $\lambda$}

\proof 
Let $g_1=g(y_{\lambda_1}),h_1=h(y_{\lambda_1}), g_2=g(y_{\lambda_2}),$
and $h_2=h(y_{\lambda_2}).$
\begin{align*}
&h_1+\lambda_1g_1\ge h_2+\lambda_1 g_2, 
\quad h_2+\lambda_2g_2\ge h_1+\lambda_2 g_1 \\
&\Leftrightarrow
 h_1-h_2+\lambda_1(g_1-g_2) \ge 0,-h_1+h_2+\lambda_2(g_2-g_1) \ge 0 \\
&\Leftrightarrow
(g_2-g_1)(\lambda_2-\lambda_1)\ge 0
\end{align*}
For $h$, change the role of $g$ and $h$. \qed

\section{Improvements for the binary search}
\label{app:search_improvements}

\subsection{Early stopping}
If $L=[\lambda_m,\lambda_M]$, and both endpoints have the same label, i.e.,  $y_{\lambda_m}=y_{\lambda_M}$, then we can terminate the binary search safely because from lemma \ref{MAP_seg}, 
it follows that the solution $y_\lambda$ will not change in this segment.


\subsection{Suboptimality bound}
Let $K(\lambda)$ be the value of the $\lambda$-oracle. i.e., 
\begin{align}\label{loracle0}
K(\lambda)=\max_{y\in\mathcal{Y}} h(y)+\lambda g(y).
\end{align}

\begin{lemma}\label{lem3}
$\Phi^*$ is upper bounded by
\begin{align} \label{optimal_upper}
\Phi(y^*)\le \dfrac{K(\lambda)^2}{4\lambda}
\end{align}
\proof 
\begin{align*}
& h(y)+\lambda g(y)\le K(\lambda)\\ 
&\iff g(y)(h(y)+\lambda g(y))\le g(y)K(\lambda)\\ 
&\iff
 \Phi(y) \le g(y) K(\lambda) - \lambda g(y)^2\\&=-\lambda\left
( g(y)- \dfrac{K(\lambda)}{2\lambda}\right )^2 + \dfrac{K(\lambda)^2}{4\lambda}\le \dfrac{K(\lambda)^2}{4\lambda}
\end{align*}
\qed
\end{lemma}

\section{Proof of the limitation of the $\lambda$-oracle search}\label{lambda_oracle_proof}

\begin{reptheorem}{loracle_l1}
  Let $\hat{H}=\max_y h(y)$ and $\hat{G}=\max_y g(y)$. For any
  $\epsilon>0$, there exists a problem with 3 labels
  such that for any $\lambda\geq0$, $\ y_\lambda=\argmin_{y\in \mathcal{Y}} \Phi(y)<\epsilon$, while
  $\Phi(y^*)=
  \dfrac{1}{4}\hat{H}\hat{G}$.Let $\hat{H}=\max_y h(y)$ and $\hat{G}=\max_y g(y)$. For any $\epsilon>0$ and $\lambda>0$, there exists a problem of 3 labels that $\ y_\lambda=\argmin_{y\in \mathcal{Y}} \Phi(y)<\epsilon$, and $\Phi(y^*)-\Phi(y_\lambda)= \dfrac{1}{4}\hat{H}\hat{G}$. 
\end{reptheorem}

\proof

We will first prove following lemma which will be used in the proof.

\begin{lemma}\label{limit_oracle}
 Let $A=[A_1\;A_2]\in \mathbb{R}^2,B=[B_1 \;B_2]\in \mathbb{R}^2$, and $C=[C_1 \;C_2]\in \mathbb{R}^2$, and $A_1<B_1<C_1$. If $B$ is under the
 line $\overline{AC}$, i.e.,$\exists t$,$0\le  t\le 1$,$D=tA+(1-t)C$, $D_1=B_1$, $D_2>B_2$. Then,   $\nexists \lambda\ge0$, $v=[1 \;\lambda]\in \mathbb{R}^2$,  such that
 \begin{align}
   v\cdot B>v\cdot A \;\text{  and }
  v\cdot B>v\cdot C \label{eq:argmaxv}
 \end{align}
 
 \proof Translate  vectors $A,B,$ and $C$ into coordinates of $[0,A_{2}],[a,b],$
 $[C_{1},0]$ by adding a vector $[-A_1,-C_2]$ to each vectors $A,B,$ and $C$,  since it does not change $B-A$ or $B-C$. Let $X=C_1$ and $Y=A_2$. 

If $0\le\lambda \le \dfrac{X}{Y}$, then $v\cdot A=\lambda Y\le X= v\cdot C$. $v\cdot (B-C)>0\iff (a-X)+\lambda b>0$ corresponds to  all the points above line  $\overline{AC}$. Similarly,
if $\lambda \ge \dfrac{X}{Y}, $ \eqref{eq:argmaxv} corresponds
to $a+\lambda (b-Y)>0$ is also all the points above  $\overline{AC}$.
\qed 
 \end{lemma}
 From lemma \ref{limit_oracle}, 
 if $y_1$,$y_2\in \mathcal{Y}$, then all the labels
 which lies under line $y_1$ and $y_2$ will not be found by $\lambda$-oracle. In the adversarial case, this holds when label lies on the line also. Therefore,
Theorem \ref{loracle_l1} holds when there exists three labels, for arbitrary small $\epsilon>0$,
$A=[\epsilon, \hat{G}],B=[\hat{H}, \epsilon]$, and $C=[\frac{1}{2}\hat{H},\frac{1}{2}\hat{G}]$, $\mathcal{Y}=\{A,B,C\}$. In this case $\hat{\Phi}\approx0$.
\qed

Therefore,
corollary \ref{cor:search_limit} holds for any problems with any $\mathcal{Y}$ that $\{A,B,C\}\subseteq\mathcal{Y}$ and $\forall y\in\mathcal{Y}$, $y$ is on or below $\overline{AC}$.

\section{Angular search}\label{angular_search_proof}
We first introduce needed notations. 
$\partial^{\bot}(a)$ be the perpendicular  slope of $a$, i.e., $\partial^{\bot}(a)=-\frac{1}{\partial(a)}=-\frac{a_1}{a_2}$. For $\mathcal{A}\subseteq\reals^2$, let label set restricted to $A$ as $\vec{\mathcal{Y}}_A=\vec{\mathcal{Y}}\cap A,  $ and 
$y_{\lambda,A}=\mathcal{O}(\lambda,A)=\argmax_{y\in\mathcal{Y}, \vec{y}\in A} h(y)+\lambda g(y)$
$=\argmax_{\vec{y}\in\vec{\mathcal{Y}}_A} [\vec{y}]_1+\lambda [\vec{y}]_2$.
Note that if $A=\reals^2,$ $y_{\lambda,\reals^2}=y_\lambda$.
For $P,Q\in \reals^2$, define $\Lambda(P,Q)$ to be the area below the line $\overline{PQ}$, i.e., $\Lambda(P,Q)=\{\vec{y}\in\reals^2|[\vec{y}]_{2}-[P]_2\le\partial^{\bot}(Q-P)([\vec{y}]_{2}-[P]_2)\}$.
  $\Upsilon_\lambda=\{\vec{y}\in\reals^2| \vec{\Phi}(\vec{y})= [\vec{y}]_{1}\cdot [\vec{y}]_{2}\ge\vec{\Phi}(\vec{y}_{\lambda,A})\}$ be the area above $C_\lambda$, and $\underline{\Upsilon}_\lambda=\{\vec{y}\in\reals^2| \vec{\Phi}(\vec{y})= [\vec{y}]_{1}\cdot [\vec{y}]_{2}\le\vec{\Phi}(\vec{y}_{\lambda,A})\}$ be the area below $C_\lambda$.
 
Recall the \emph{constrained $\lambda$-oracle}  defined in \eqref{eq:constrained_oracle}:
\begin{align*}
  y_{\lambda,\alpha,\beta}=\mathcal{O}_{c}(\lambda,\alpha,\beta)=\max_{y\in\mathcal{Y},\; \alpha h(y)\ge g(y),\; \beta h(y)<g(y)} \mathcal{L}_\lambda(y)  
\end{align*}
where $\alpha,\beta\in\reals_+$ and $\alpha\ge\beta>0$. Let $A(\alpha,\beta)\subseteq\reals^2$ be the restricted search space, i.e., $A(\alpha,\beta)=\{a\in\reals^2|\beta<\partial(a)\le\alpha\}$. Constrained $\lambda$-oracle reveals maximal $\mathcal{L}_\lambda$ label within restricted area defined by $\alpha$ and $\beta$. The area is bounded by two lines whose slope is $\alpha$ and $\beta$. 
 Define a pair $(\alpha,\beta),\alpha ,\beta\in \reals_+,\alpha\ge\beta>0$ as an {\em angle}.  The angular search recursively divides an angle into two different angles, which we call the procedure as a {\em split}. For $\alpha\ge\beta\ge0$, let $\lambda=\dfrac{1}{\sqrt{\alpha\beta}}$, $z=\vec{y}_{\lambda,\alpha,\beta}$ and $z'=[\lambda [z]_2,\frac{1}{\lambda}[z]_1]$.  Let $P$ be the point among $z$ and $z'$ which has the greater slope (any if two equal), and $Q$ be the other point, i.e., if $\partial(z)\ge \partial(z')$, $P=z$ and $Q=z'$, otherwise $P=z'$ and $Q=z$.  Let $R=\left [\sqrt{\lambda[z]_1\cdot [z]_2}\;\; \sqrt{\frac{1}{\lambda}[z]_1\cdot [z]_2}\;\right ]$. Define split$(\alpha,\beta)$ as a procedure divides $(\alpha,\beta)$ 
into two angles $(\alpha^+,\gamma^+)=(\partial(P),\partial(R))$
and  $(\gamma^+,\beta^+)=(\partial(R),\partial(Q))$.

First, show that $\partial(P)$ and $\partial(Q)$ are in between $\alpha$ and $\beta$, and $\partial(R)$ is  between $\partial(P)$ and $\partial(Q)$. 
\begin{lemma} \label{lemma:angle_range}For each split$(\alpha,\beta)$,
\begin{align*}
\beta\le&\partial(Q)\le\partial(R)\le\partial(P)\le\alpha 
\end{align*}
\proof
$\beta\le\partial(z)\le\alpha$ follows from the definition of constrained $\lambda$-oracle in \eqref{eq:constrained_oracle}.

$\partial(z')=\dfrac{1}{\lambda^2\partial(z)}=\dfrac{\alpha\beta}{\partial(z)}\implies \beta\le\partial(z')\le\alpha\implies\beta\le\partial(Q)\le\partial(P)\le\alpha$.

$\partial(Q)\le\partial(R)\le\partial(P)\iff$
 $\min\left \{\partial(z),\dfrac{1}{\lambda^2\partial(z)}\right \}\le \dfrac{1}{\lambda}\le\max\left \{\partial(z),\dfrac{1}{\lambda^2\partial(z)}\right \}$ from $\forall a,b\in \reals_+,b\le a\implies  b\le\sqrt{ab}\le a$. \qed
\end{lemma}
After each split, the union of the divided angle ($\alpha^+,\gamma$) and ($\gamma,\beta^{+}$) can be smaller than angle $(\alpha,\beta)$. However, following lemma shows it is safe to use ($\alpha^+,\gamma$) and ($\gamma,\beta^{+}$) when our objective is to find $y^*$.
\begin{lemma} \label{lem:shrink_angel}
\begin{align*}
\forall a\in\vec{\mathcal{Y}}_{A(\alpha,\beta)}, \Phi(a)> \Phi(y_{\lambda,\alpha,\beta}) \implies \beta^+<\partial(a)<\alpha^+ 
\end{align*}
\proof From lemma \ref{lem:restricted_space}, $\vec{\mathcal{Y}}_{A(\alpha,\beta)}\subseteq\Lambda(P,Q)$. Let $U=\{a\in\reals^2|\partial(a)\ge \alpha_+=\partial(P)\}$,  $B=\{a\in\reals^2|\partial(a)\le \beta_+=\partial(Q)\}$,  and two contours of function $C=\{a\in\reals^2|\vec{\Phi}(a)= \Phi(y_{\lambda,\alpha,\beta})\}$,  $S=\{a\in\reals^2|\mathcal{L}_\lambda(a)=\mathcal{L}_\lambda(\vec{y}_{\lambda,\alpha,\beta})\}$. $S$ is the upper bound of $\Lambda(P,Q)$, and  $C$ is the upper bound of $\underline{C}=\{a\in \reals^2|\vec{\Phi}(a)\le\Phi(y_{\lambda,\alpha,\beta})\}$. $P$ and $Q$ are the intersections of $C$ and $S$. For area of $U$ and $B$, $S$ is under $C$, therefore, $\Lambda(P,Q)\cap U\subseteq \underline{C} $, and $\Lambda(P,Q)\cap B\subseteq \underline{C}$. It implies that $\forall a\in (\Lambda(P,Q)\cap U )\cup (\Lambda(P,Q)\cap B) \implies \vec{\Phi}(a)\le \Phi(y_{\lambda,\alpha,\beta})$. And the lemma follows from $A(\alpha,\beta)=U\cup B\cup \{a\in \reals^2| \beta^+<\partial(a)<\alpha^+\}$.\qed
\end{lemma}

We  associate a quantity we call a {\em capacity} of an angle, which is used to prove the suboptimality of the algorithm. For an angle $(\alpha,\beta)$, the capacity of an angle $v(\alpha,\beta)$ is
\begin{align*}
  v(\alpha,\beta):=\sqrt{\frac{\alpha}{\beta}}
\end{align*}
Note that from the definition of an angle, $v(\alpha,\beta)\ge1$. First show that the capacity of angle decreases exponentially for each split. 

\begin{lemma} \label{lem_split_decrease}
 For any angle $(\alpha,\beta)$ and its split $(\alpha^+,\gamma^+)$ and  $(\gamma^+,\beta^+)$, 
\begin{align*}
 v(\alpha,\beta)\ge v (\alpha^+,\beta^+)=v(\alpha^+,\gamma^+)^{2}=v(\gamma^+,\beta^+)^{2}
\end{align*}
\proof Assume $\partial(P)\ge \partial(Q)$ (the other case is follows the same proof with changing the role of $P$ and $Q$), then $\alpha^+=\partial(P)$ and $\beta^+=\partial(Q)$. $\partial(Q)=\dfrac{1}{\lambda^2\partial(P)}=\dfrac{\alpha\beta}{\partial(P)},v$ $(\alpha^+,\beta^+)=v(\partial(P),\partial(Q))=\lambda \partial(P)=\dfrac{\partial(P)}{\sqrt{\alpha\beta}}.$ Since $\alpha$ is the upper bound and $\beta$ is the lower bound of $\partial(P)$,  $\sqrt{\dfrac{\beta}{\alpha}} \le v(\partial(P),\partial(Q))\le \sqrt{\dfrac{\alpha}{\beta}}$.  Last two equalities in the lemma are from $v(\partial(P),\partial(R))=v(\partial(R),\partial(Q))=\sqrt{\frac{\partial(P)}{\sqrt{\alpha\beta}}}$  by plugging in the coordinate of $R$.\qed
 \end{lemma}
\begin{lemma}\label{lem:suboptimal__} Let $\mathcal{B}(a)=  \dfrac{1}{4}\left(a+\dfrac{1}{a}\right)^2$ . The suboptimality bound of an angle $(\alpha,\beta)$ with $\lambda=\dfrac{1}{\sqrt{\alpha\beta}}$ is 
\begin{align*} 
\dfrac{\max_{\vec{y}\in \vec{\mathcal{Y}}_{A(\alpha,\beta)}}\vec{\Phi}(\vec{y})}{\Phi(y_{\lambda,\alpha,\beta})}
\le 
 \mathcal{B}(v(\alpha,\beta)).
\end{align*}
\proof From lemma \ref{lem:restricted_space}, $\vec{\mathcal{Y}}_{A(\alpha,\beta)}\subseteq\Lambda(P,Q)=\Lambda(z,z').$  Let $\partial(z)=\gamma$. From \ref{lemma:angle_range}, $\beta\le\gamma\le \alpha$.
Let  $m=\argmax_{a\in \Lambda(z,z')}\vec{\Phi}(a)$. $m$ is on line $\overline{zz'}$ otherwise we can move $m$ increasing direction of each axis till it meets the boundary $\overline{zz'}$ and  $\Phi$ only increases, thus $m=tz+(1-t)z'$. $\vec{\Phi}(m)=\max_{t}\vec{\Phi}(tz+(1-t)z')$. $\dfrac{\partial\vec{\Phi}(tz+(1-t)z')}{\partial t}=0 \implies t=\dfrac{1}{2}$. $m=\frac{1}{2}[z_1+\lambda z_2\;\; z_2+\frac{z_1}{\lambda}]$.
\begin{align*}
\dfrac{\max_{\vec{y}\in \vec{\mathcal{Y}}_{A(\alpha,\beta)}}\vec{\Phi}(\vec{y})}{\Phi(y_{\lambda,\alpha,\beta})}=\dfrac{1}{4}\left (\sqrt{\dfrac{z_1}{\lambda z_2}}+\sqrt{\dfrac{\lambda z_2}{z_1}}\right)^2\\=\dfrac{1}{4}\left (\sqrt{\dfrac{\sqrt{\alpha\beta} }{\gamma}}+\sqrt{\dfrac{\gamma}{\sqrt{\alpha\beta} }}\right)^2
\end{align*} 
Since $v(a)=v\left(\frac{1}{a}\right)$ and $v(a)$ increases monotonically for $a\ge1$,
\begin{align*}
\mathcal{B}(a)\le \mathcal{B}(b)\iff \max\left \{a,\frac{1}{a}\right\} \le \max\left \{b,\frac{1}{b}\right\}
\end{align*}
If $\dfrac{\sqrt{\alpha\beta} }{\gamma}\ge \dfrac{\gamma}{\sqrt{\alpha\beta} }$, then $\dfrac{\sqrt{\alpha\beta} }{\gamma}\le \sqrt{\dfrac{\alpha}{\beta}}$ since $\gamma\ge \beta$. If $\dfrac{\gamma}{\sqrt{\alpha\beta}}\ge \dfrac{\sqrt{\alpha\beta} }{\gamma}$, then $\dfrac{\gamma}{\sqrt{\alpha\beta}}\le \sqrt{\dfrac{\alpha}{\beta}}$ since $\gamma\le \alpha$. Therefore, $\dfrac{\max_{\vec{y}\in \vec{\mathcal{Y}}_{A(\alpha,\beta)}}\vec{\Phi}(\vec{y})}{\Phi(y_{\lambda,\alpha,\beta})}
=\mathcal{B}\left (\dfrac{\sqrt{\alpha\beta} }{\gamma}\right)\le \mathcal{B}(v(\alpha,\beta))$.\qed
\end{lemma}

 Now we can prove the theorems.

\begin{reptheorem}{thm:angular_optimal} Angular search described in algorithm \ref{alg:angular} finds optimum $y^*=\argmax_{y\in \mathcal{Y}} \Phi(y) $ at most $t=2M+1$ iteration where $M$ is the number of the labels.
\proof
 Denote $y_{t}, \alpha_t, \beta_t, z_t,z_t', K_t^1,$ and $K_t^2$ for $y,\alpha,\beta,z,z',K^1,$ and $K^2$  at iteration $t$ respectively.  $\mathcal{A}(\alpha_t,\beta_t)$ is the search space at each iteration $t$. At the first iteration $t=1$, the search space contains all the labels with positive $\Phi$, i.e., $\{y|\Phi(y)\ge 0 \}\subseteq\mathcal{A}(\infty,0)$. At iteration $t>1$, firstly, when $y_t=\emptyset$,  the search area $\mathcal{A}(\alpha_t,\beta_t)$ is removed from the search since  $y_t=\emptyset$ implies there is no label inside $\mathcal{A}(\alpha_t,\beta_t)$. Secondly, when $y_t\neq\emptyset$, $\mathcal{A}(\alpha_t,\beta_t)$ is dequeued, and $K_t^1$ and $K_t^2$ is enqueued. 
From lemma \ref{lem:shrink_angel}, at every step, we are ensured that do not loose $y^*$. 
  By using strict inequalities in the constrained oracle with valuable $s$, we can ensure $y_t$ which oracle returns is an unseen label. Note that split only happens if a label is found, i.e., $y_t\ne\emptyset$. Therefore, there can be only $M$ splits, and each split can be viewed as a branch in the binary tree, and the number of queries are the number of nodes. Maximum number of the nodes with $M$ branches are $2M+1$. \qed
\end{reptheorem}

\begin{reptheorem}{thm:angular_suboptimality} In angular search described in algorithm \ref{alg:angular}, at iteration $t$, 
\begin{align*}
\dfrac{\Phi(y^*)}{\Phi(\hat{y})}\le (v_1)^{\frac{4}{t+1}}
\end{align*}
 where $v_1=\max \left \{\dfrac{\lambda_0}{\partial(y_1)}, \dfrac{\partial(y_1)}{\lambda_0}\right\}$, $\lambda_0$ is the initial $\lambda$ used, and $y_1$ is the first label returned by constrained $\lambda$-oracle.
\proof
After $t\ge2^{r}-1$ iteration as in algorithm \ref{alg:angular} where $r$ is an integer, for all the angle $(\alpha,\beta)$ in the queue $Q$, $v(\alpha,\beta)\le (v_1)^{2^{1-r}}$. 
This follows from the fact that since the algorithm uses the depth first search, after $2^{r}-1$ iterations all the nodes at the search is at least $r$. At each iteration, for a angle, the capacity is square rooted from the lemma \ref{lem_split_decrease}, and the depth is increased by one. 
And the theorem follows from the fact that after $t\ge 2^{r}-1$ iterations, all splits are at depth $r'\ge r$, and at least one of the split contains the optimum with suboptimality bound with lemma \ref{lem:suboptimal__}. Thus,
\begin{align*}
\dfrac{\Phi(y^*)}{\Phi(\hat{y})}\le \mathcal{B}\left ((v_1)^{2^{1-r}} \right ) < (v_1)^{2^{2-r}} \le(v_1)^{\frac{4}{t+1}}
\end{align*}
\qed
\end{reptheorem}
\begin{reptheorem}{thm:angular_running_time}  Assuming $\Phi(y^*)>\phi$, angular search described in algorithm \ref{alg:angular} with $\lambda_0=\dfrac{\hat{G}}{\hat{H}}, \alpha_0=\dfrac{\hat{G}^2}{\phi},\beta_0=\dfrac{\phi}{\hat{H}^2}$, finds $\epsilon$-optimal solution, $\Phi(y)\ge(1-\epsilon)\Phi(y^*)$, in $T$ queries and $O(T)$ operations where $T=4\log\left(\dfrac{\hat{G}\hat{H}}{\phi}\right)\cdot \dfrac{1}{\epsilon}$, and $\delta$-optimal solution, $\Phi(y)\ge\Phi(y^*)-\delta$,  in $T'$ queries and $O(T')$ operations where $T'= 4\log\left(\dfrac{\hat{G}\hat{H}}{\phi}\right)\cdot \dfrac{\Phi(y^*)}{\delta}$.
\proof $\Phi(y^*)>\phi\Leftrightarrow  \dfrac{\phi}{\hat{H}^2}  <\dfrac{g(y^*)}{h(y^*)}=\partial(\vec{y^*})< \dfrac{\hat{G}^2}{\phi}$. $v_1=\max \left \{\dfrac{\lambda_0}{\partial(y_1)}, \dfrac{\partial(y_1)}{\lambda_0}\right\}$ from Theorem \ref{thm:angular_suboptimality}. Algorithm finds $y^*$ if $\beta\le \partial(\vec{y^*})\le \alpha$, thus set  $\alpha= \dfrac{\hat{G}^2}{\phi}$ and $\beta=\dfrac{\phi}{\hat{H}^2}$. Also from the definition of constrained $\lambda$-oracle, $\beta=\dfrac{\phi}{\hat{H}^2}\le\partial(y_1)\le\alpha= \dfrac{\hat{G}^2}{\phi}$. Therefore, $v_1\le\max \left \{\dfrac{\lambda_0}{\partial(y_1)}, \dfrac{\partial(y_1)}{\lambda_0}\right\}$. And the upper bound of two terms equal when $\lambda_0=\dfrac{\hat{G}}{\hat{H}}$, then $v_1\le  \dfrac{\hat{G}\hat{H}}{\phi}$. $\delta$ bound follows plugging in the upper bound of $v_1$, and $\epsilon=\dfrac{\delta}{\Phi(y^*)}$. \qed
\end{reptheorem}


 \begin{algorithm}
  \caption{Angular search }
\label{alg:angular}
  \begin{algorithmic}[1]
    \Procedure{AngularSearch}{$\lambda_0,T$}
    \INPUT {$\lambda_{0}\in\mathbb{R}_+$, 
       and maximum iteration $T\in \reals_{+}$
}   \OUTPUT{$\hat{y} \in \mathcal{Y}.$}
    \INIT  {$\alpha_0=\infty,\beta_0=0,$ Empty queue $\mathcal{Q}$, $\hat{y}=\emptyset$.$\lambda\gets \lambda_0$}
    \State $\text{ADD}(\mathcal{Q},(\alpha,\beta,0))$
    \While{$\mathcal{Q}\ne\emptyset$} 
    \State $(\alpha,\beta,s)\gets \text{Dequeue}(\mathcal{Q})$
    \If {$\beta\ne0$}
        \State $\lambda\gets \frac{1}{\sqrt{\alpha\beta}}$ 
    \EndIf
    \If {$s=0$}
        \State $y\gets\mathcal{O}_c(\lambda,\alpha,\beta)$   
    \Else
        \State $y\gets\underline{\mathcal{O}}_c(\lambda,\alpha,\beta)$   
    \EndIf
    \If {$\Phi(y)> \Phi(\hat{y})$}
    \State $\hat{y}\gets y$
    \EndIf
    \If {$y\ne\emptyset$}
    \State $z\gets [h(y)\; g(y)],z'\gets [\lambda g(y)\; \dfrac{1}{\lambda}h(y)]$
    \State $r\gets \left [\sqrt{\lambda h(y) g(y)}\; \sqrt{\frac{1}{\lambda} h(y) g(y)}\; \right ]$
    \If {$z_1=z'_1$}
    \State return $y$
    \ElsIf {$\partial(z)>\partial(z')$}
     \State $K^{1}\gets(\partial(z),\partial(r),1)$
     \State $K^{2}\gets(\partial(r),\partial(z'),0)$
    \Else
     \State $K^{1}\gets(\partial(z'),\partial(r),1)$
     \State $K^{2}\gets(\partial(r),\partial(z),0)$
    \EndIf
     \State $\text{ADD}(\mathcal{Q},K^1)$ $.\text{ADD}(\mathcal{Q},K^2)$ 
    \EndIf
    \State $t\gets t+1$
    \If {$t=T$} \Comment{maximum iteration reached}
    \State \Return $\hat{y}$ 
    \EndIf
    \EndWhile    
    \EndProcedure
  \end{algorithmic}
\end{algorithm}

\section{Illustration of the angular search}\label{ap:illu}
Following figure \ref{fig:angular_illu} illustrates Angular search.
Block dots are the labels from figure \ref{fig:all_points}. Blue X denotes the new label returned by the oracle. Red X is the maximum point. Two straights lines are the upper bound and the lower bound used by the constrained oracle. Constrained oracle returns a blue dot between the upper and lower bounds. We can draw a line that passes blue X that no label can be above the line.  Then, split the angle into half. This process continues until the $y^*$ is found.   
\begin{figure}[H]
\centering     
\begin{subfigure}[b]{\linewidth}
\includegraphics[width=\linewidth]{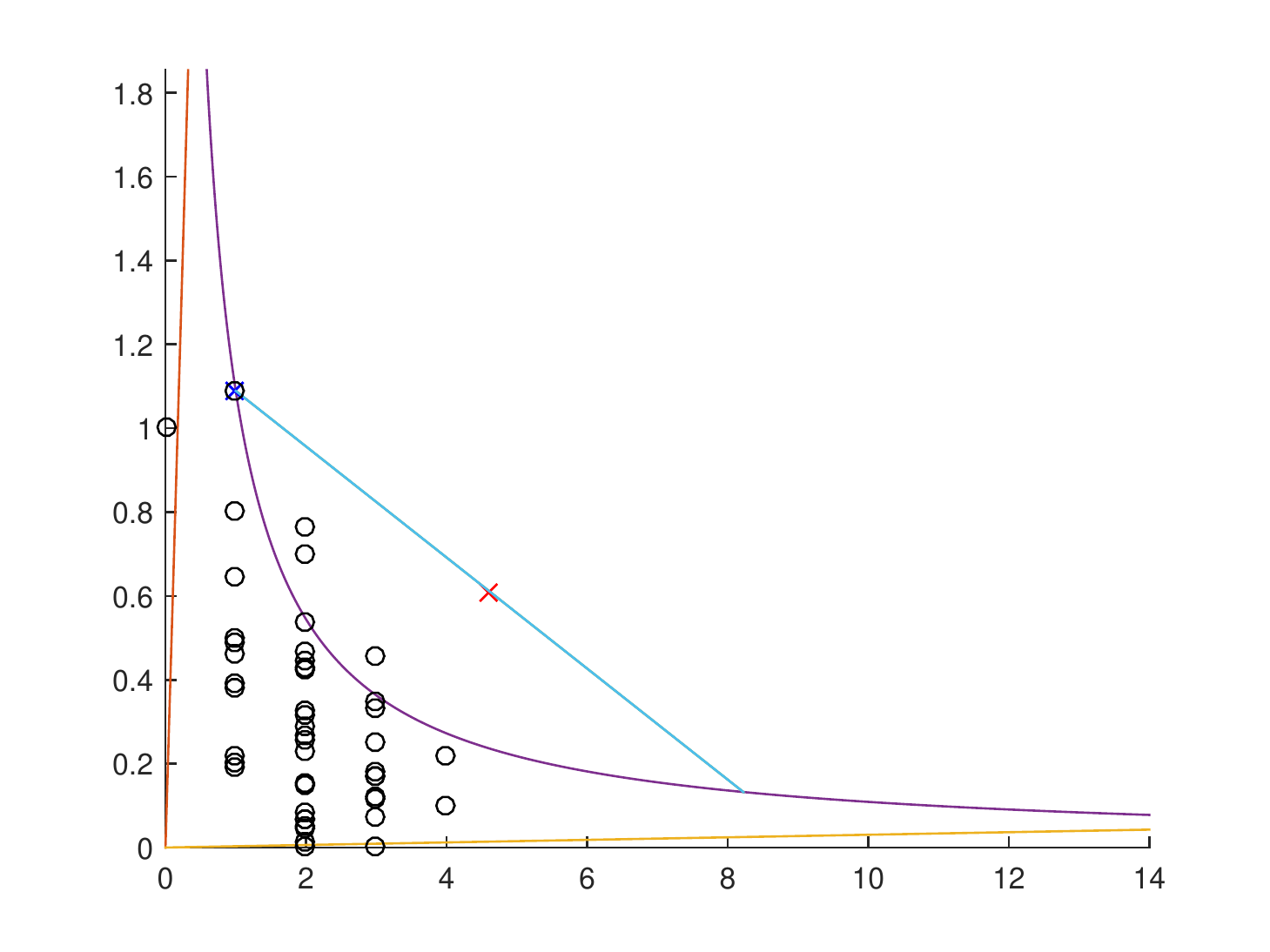}
\caption{Iteration 1}
\end{subfigure}
\begin{subfigure}[b]{\linewidth}
\includegraphics[width=\linewidth]{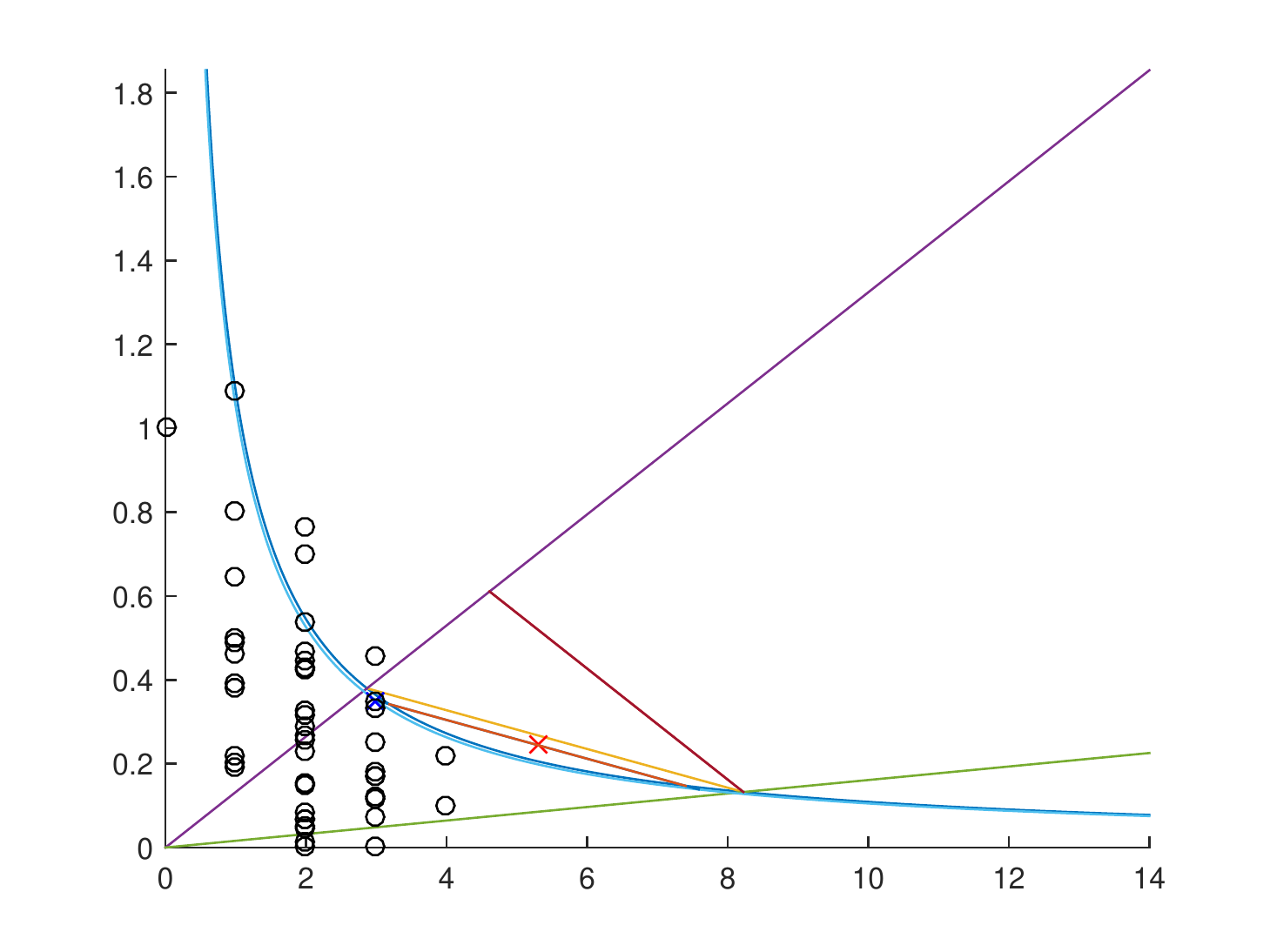}
\caption{Iteration 2}
\end{subfigure}
\begin{subfigure}[b]{\linewidth}
\includegraphics[width=\linewidth]{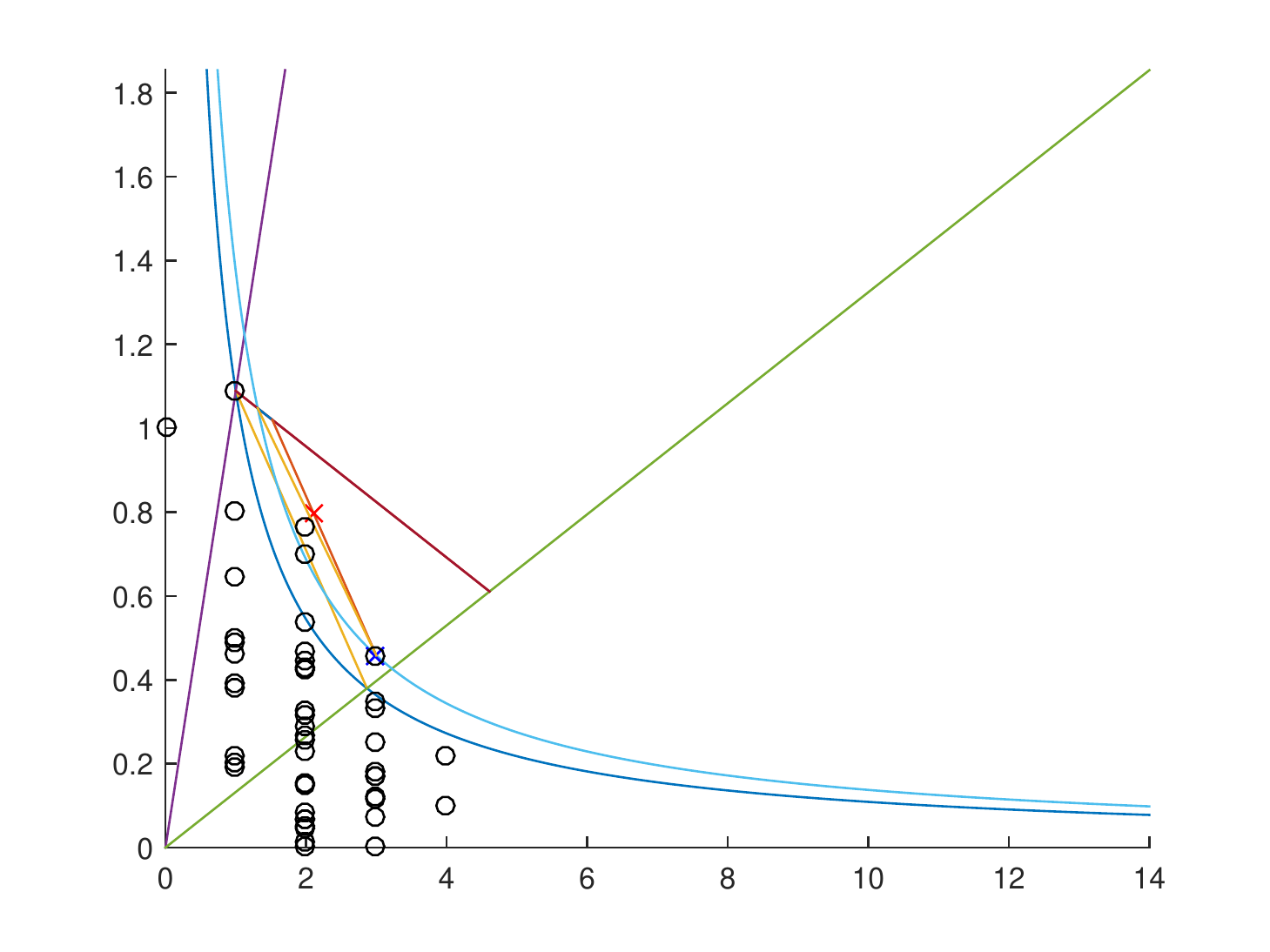}
\caption{Iteration 3}
\end{subfigure}
\caption{Illustration of the Angular search.}\label{fig:angular_illu}
\end{figure}

\section{Limitation of the constraint $\lambda$-oracle search}
\begin{reptheorem}{loracle_l2}
Any search algorithm accessing labels only through   $\lambda$-oracle with any number of the linear constraints cannot find $y^*$ in less than $M$ iterations in the worst case where $M$ is the number of labels.
\proof
 We show this in the perspective of a game between a searcher and an oracle. At each iteration, the searcher query the oracle with $\lambda$ and the search space denoted as $\mathcal{A}$, and the oracle reveals a label according to the query. And the claim is that with any choice of $M-1$ queries, for each query the oracle can either give an consistent label or indicate that there is no label in $\mathcal{A}$ such that after $M-1$ queries the oracle provides an unseen label $y^*$ which has bigger $\Phi$ than all previous revealed labels.

 Denote each query at iteration $t$ with $\lambda_t>0$ and a query closed and convex set $\mathcal{A}_t\subseteq\reals^2$
, and denote the revealed label at iteration $t$ as $y_t$. We will use   $y_t=\emptyset$ to denote that there is no label inside query space $\mathcal{A}_t$. Let $\mathcal{Y}_t=\{y_{t'}|t'<t\}$.

 Algorithm \ref{alg:M_iter} describes the pseudo code for  generating such $y_t$. The core of the algorithm is maintaining a rectangular area $\mathcal{R}_t$ for each iteration $t$ with following properties. Last two properties are for $y_t$. 
%
%
\begin{enumerate}
 \item $\forall t'<t,\forall y\in \mathcal{R}_t, \Phi(y)>\Phi(y_{t'})$.
 \item $\forall t'<t,\forall y\in \mathcal{R}_t\cap \mathcal{A}_{t'}, h(y_{t'})+\lambda_{t'} g(y_{t'})> h(y)+\lambda_{t'} g(y)$.
 \item  $\mathcal{R}_t\subseteq \mathcal{R}_{t-1}$.
 \item $\mathcal{R}_t$ is a non-empty open set.
 \item $y_t\in \mathcal{R}_t\cap \mathcal{A}_t$
 \item $y_t=\argmax_{y\in \mathcal{Y}_{t}\cap \mathcal{A}_t} h(y)+\lambda_{t} g(y)$.
 \end{enumerate}

Note that if these properties holds till iteration $M$, we can simply 
set $y^{*}$ as any label in $\mathcal{R}_M$ which proves the claim. 

First, we show that property 4 is true. $\mathcal{R}_0$ is a non-empty open set.  Consider iteration $t$, and assume $\mathcal{R}_{t-1}$ is a non-empty open set. Then $\tilde{R}$ is an open set since $\mathcal{R}_{t-1}$ is an open set.
There are two unknown functions,  $Shrink$ and $FindRect$. 
For open set $A\subseteq\reals^2,y\in \reals^2$, let $Shink(A,y,\lambda)=A-\{y'|\Phi(y')\le\Phi(y) $ or $h(y')+\lambda g(y')\ge h(y)+\lambda g(y)\}$. Note that $Shrink(A,y,\lambda)\subseteq A$, and $Shrink(A,y,\lambda)$  is an open set. Assume now that there exists a $y$ such that $Shrink(\mathcal{R}_{t-1},y,\lambda_t)\ne \emptyset$ and $FindPoint(\mathcal{R}_{t-1},\lambda_t)$  returns such $y$. Function $FindPoint$ will be given later. 
$FindRect(A)$ returns an open non-empty rectangle inside $A$. Note that $Rect(A)\subseteq\ A$, and since input to $Rect$ is always non empty open set, such rectangle exists. Since $\mathcal{R}_0$ is non-empty open set, $\forall t,\mathcal{R}_t$ is a non-empty open set.

Property 3 and 5 are easy to check. Property 1 and 2 follows from the fact that $\forall t\in \{t|y_t\ne \emptyset\},\forall t'>t, \mathcal{R}_{t'}\subseteq Shrink(\mathcal{R}_{t-1},y_t,\lambda_{t-1})$.

Property 6 follows from the facts that if $\mathcal{Y}_{t-1}\cap\mathcal{A}_t\neq \emptyset$, $\tilde{\mathcal{R}}=0\implies \mathcal{R}_{t-1} \subseteq \{y| h(y)+\lambda_{t} g(y)>  h(\tilde{y})+\lambda_{t} g(\tilde{y}) $ and $ y\in\mathcal{A}_t\}$, otherwise $\mathcal{Y}_{t-1}\cap\mathcal{A}_t= \emptyset$, and $\mathcal{R}_{t-1} \subseteq \mathcal{A}_t$.

\begin{algorithm}
  \caption{Construct a consistent label set $\mathcal{Y}$.   }
\label{alg:M_iter}
  \begin{algorithmic}[1]
    \INPUT {$\{ \lambda_t, \mathcal{A}_t\}_{t=1}^{M-1},  \lambda_t>0, \mathcal{A}_t\subseteq\reals^2, \mathcal{A}_t $is closed and convex region. 
\OUTPUT {$\{y_t\in\reals^2\}_{t=1}^{t=M-1},y^*\in \reals^2$} 
}
    \INIT  {$\mathcal{R}_0=\{(a,b)|0< a , 0<
  b\}$},$\mathcal{Y}_{0}=\emptyset$.
    \For {$t=1,2,\dots,M-1$ }
     \If{$ \mathcal{Y}_{t-1}\cap\mathcal{A}_t= \emptyset$}  
\State $\tilde{y}=\argmax_{y\in \mathcal{Y}_t}  h(y)+\lambda_{t} g(y).$
\State $\tilde{\mathcal{R}}=\mathcal{R}_{t-1} \cap \{y| h(y)+\lambda_{t} g(y)<  h(\tilde{y})+\lambda_{t} g(\tilde{y}) $ or $ y\notin\mathcal{A}_t\}$.
     \Else 
     \State $\tilde{y}=\emptyset$, $\tilde{\mathcal{R}}=\mathcal{R}_{t-1}-\mathcal{A}_t$.
     \EndIf
     \If{$\tilde{\mathcal{R}}\ne\emptyset$}
     \State $y_t=\emptyset$. $\mathcal{R}_t=FindRect(\tilde{\mathcal{R}})$
     \Else 
     \State $y_t=FindPoint(\mathcal{R}_{t-1},\lambda_t)$.
     \State $\mathcal{R}_t=FindRect(Shrink(\mathcal{R}_{t-1},y_t,\lambda_t)).$
     \EndIf
     \If{$y_t\ne \emptyset$}
     \State $\mathcal{Y}=\mathcal{Y}\cup\{y_t\}$.
     \EndIf
    \EndFor
    \State Pick any $y^*\in \mathcal{R}_{M-1}$
  \end{algorithmic}
\end{algorithm}

 $FindPoint(A,\lambda)$ returns any $y\in A -\{y\in \reals^2|\lambda y_2=y_1\}$. Given input $A$ is always an non-empty open set, such $y$ exists. $Shrink(\mathcal{R}_{t-1},y,\lambda_t)\ne \emptyset$ is ensured from the fact that two boundaries, $c=\{y'|\Phi(y')=\Phi(y)\}$ and $d=\{h(y')+\lambda g(y')= h(y)+\lambda g(y)\}$ meets at $y.$ Since $c$ is a convex curve, $c$ is under $d$ on one side. Therefore the intersection of set above $c$ and below $d$ is non-empty and also open.
\qed

\end{reptheorem}

\section{Additional Plots from the Experiments}\label{ap:additional_plot}

\begin{figure}[H]
\centering     
\begin{subfigure}[b]{\linewidth}
\includegraphics[width=\linewidth]{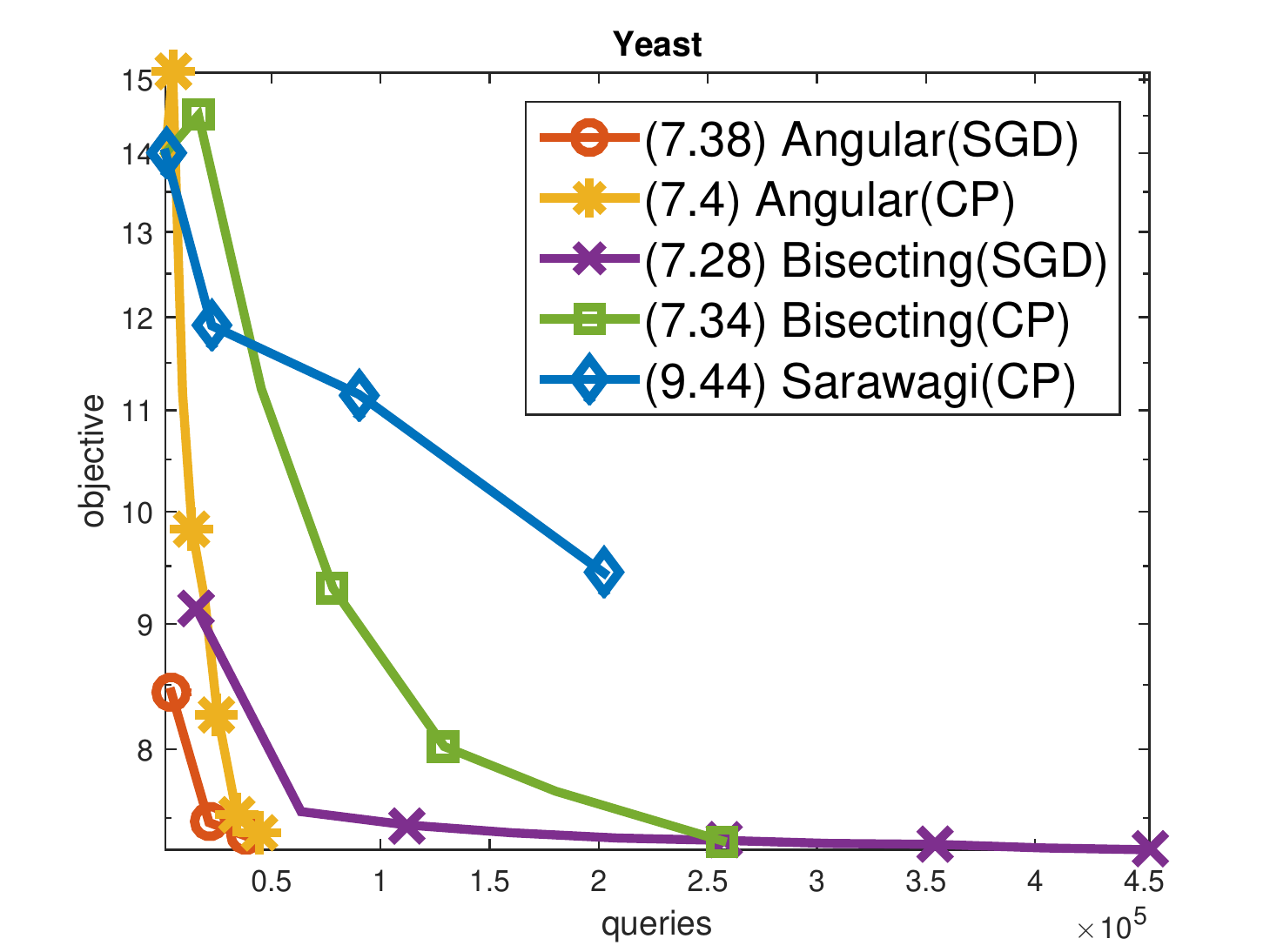}
\caption{Objective vs  queries}
\end{subfigure}
\begin{subfigure}[b]{\linewidth}
\includegraphics[width=\linewidth]{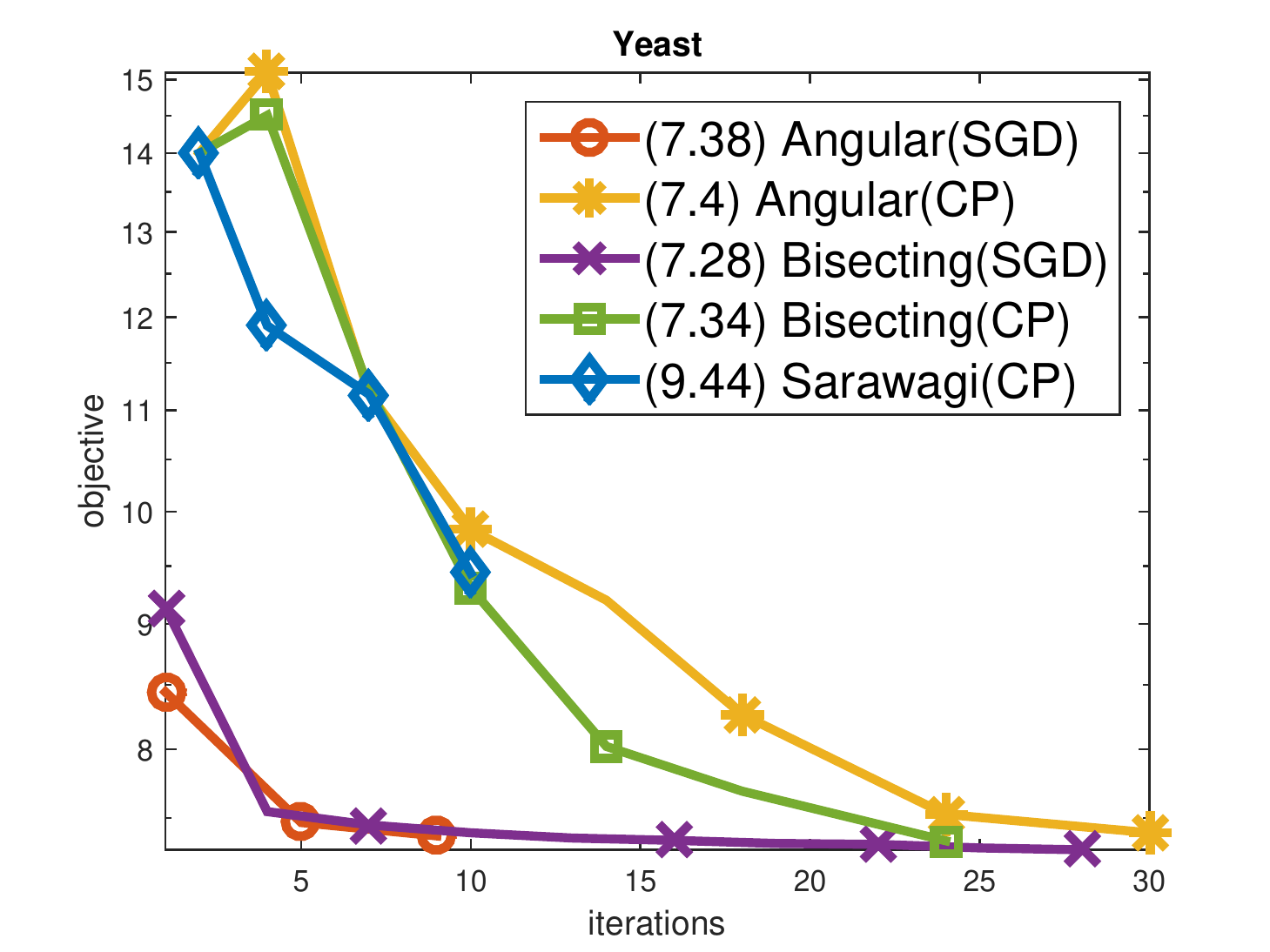}
\caption{Objective vs iterations}
\end{subfigure}
\begin{subfigure}[b]{\linewidth}
\includegraphics[width=\linewidth]{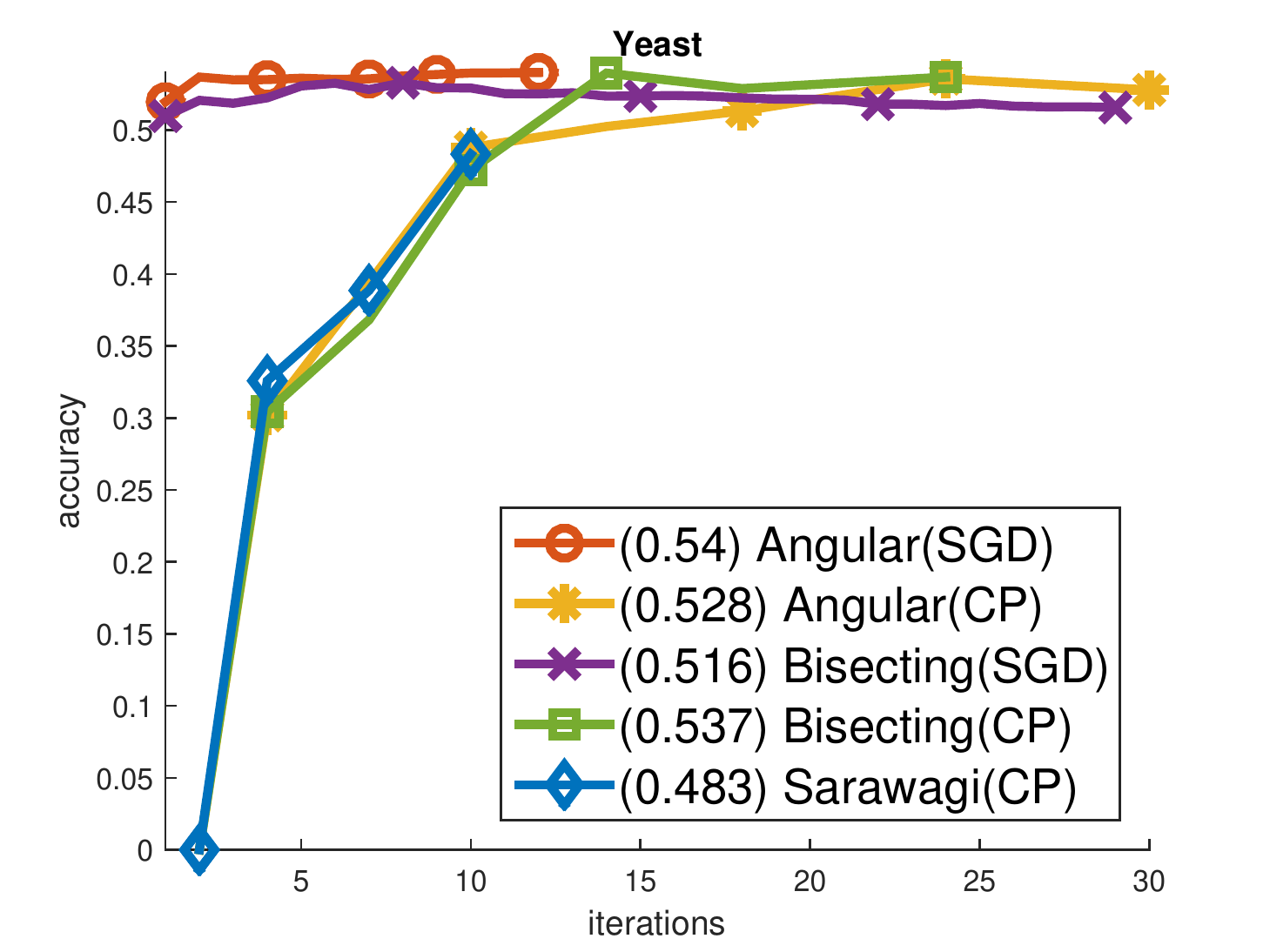}
\caption{Accuracy vs iterations}
\end{subfigure}
\end{figure}

\begin{figure}[H]
\centering     
\begin{subfigure}[b]{\linewidth}
\includegraphics[width=\linewidth]{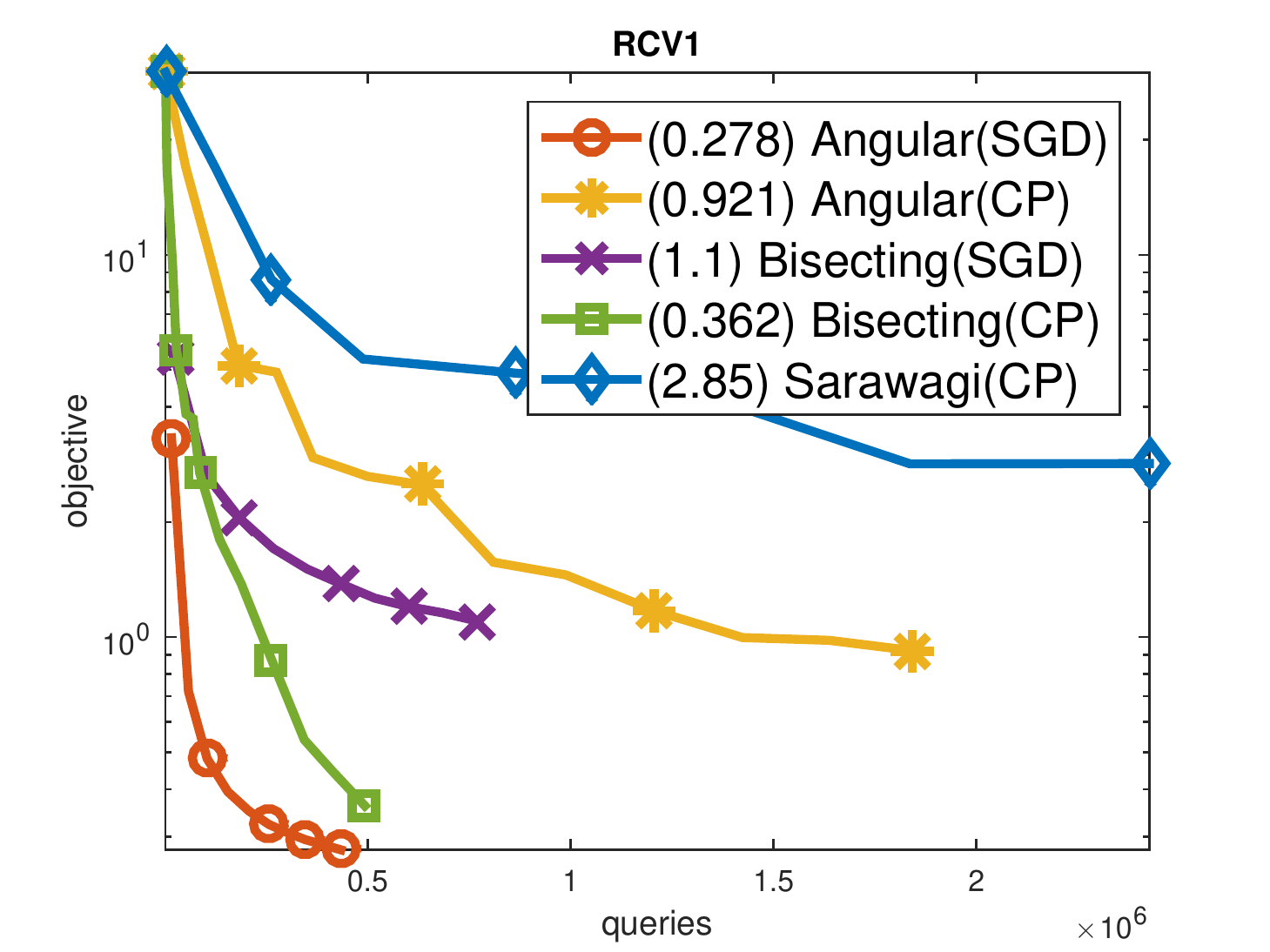}
\caption{Objective vs  queries}
\end{subfigure}
\begin{subfigure}[b]{\linewidth}
\includegraphics[width=\linewidth]{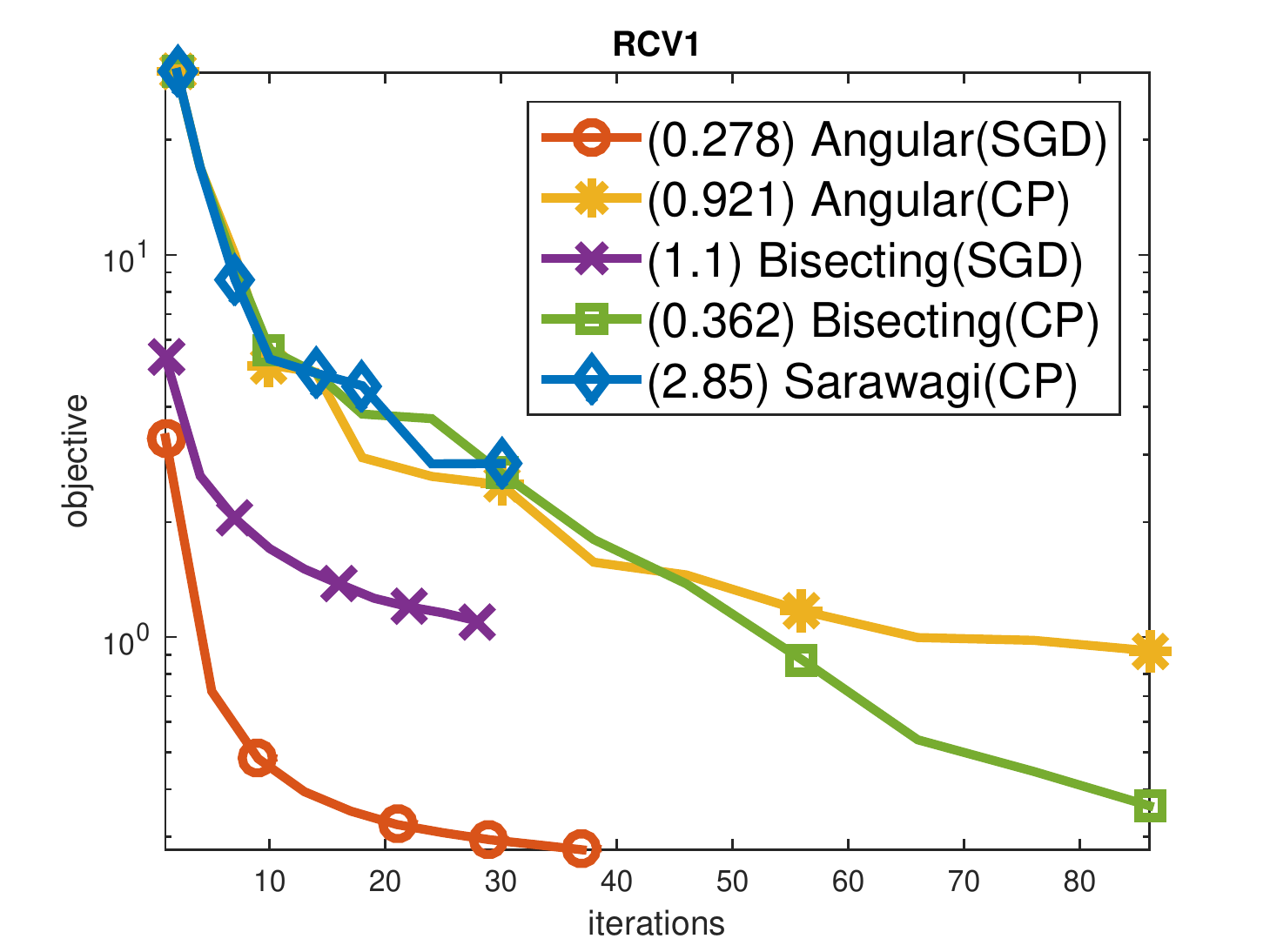}
\caption{Objective vs iterations}
\end{subfigure}
\begin{subfigure}[b]{\linewidth}
\includegraphics[width=\linewidth]{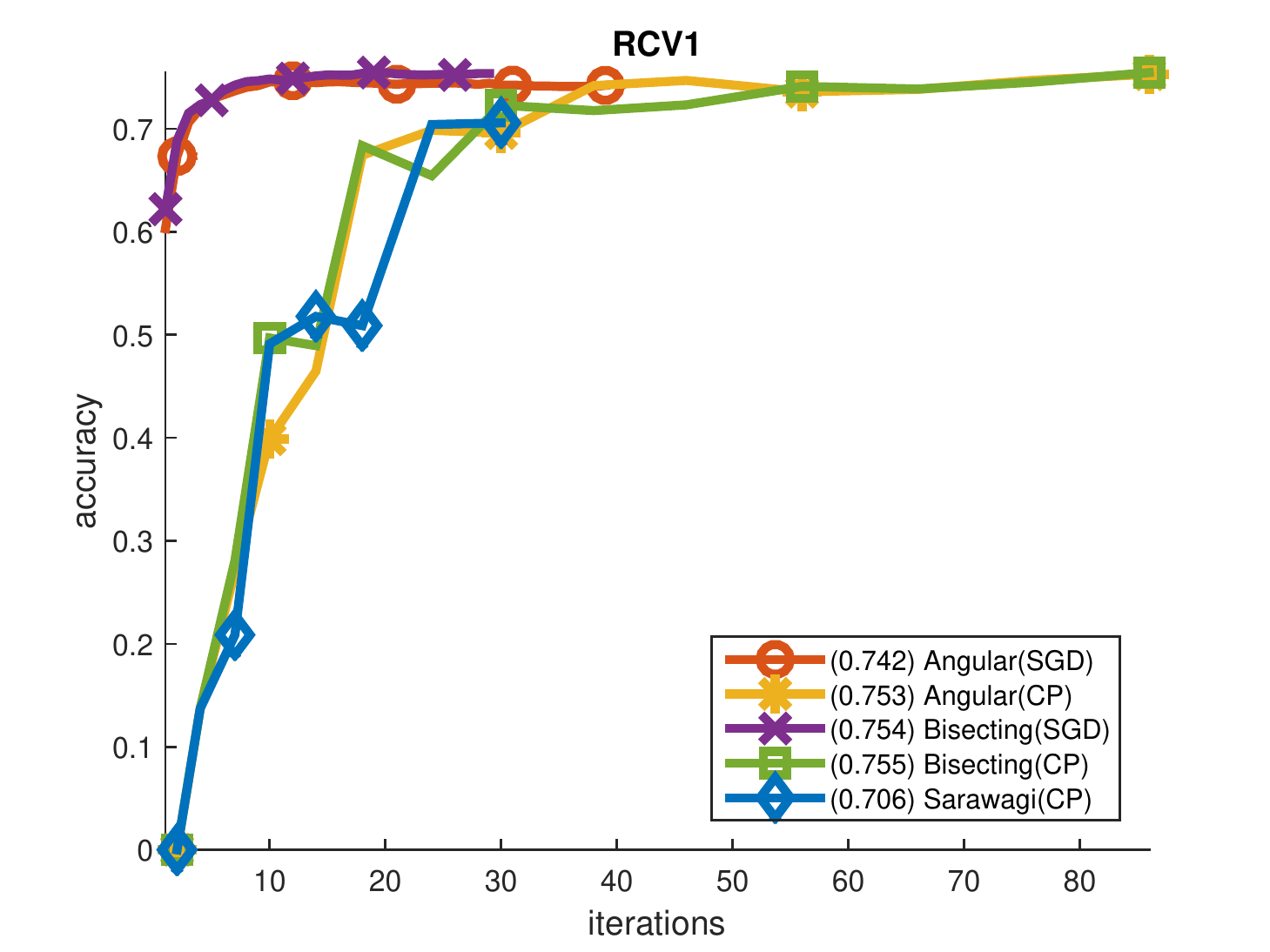}
\caption{Accuracy vs iterations}
\end{subfigure}
\caption{Additional experiment plot (RCV)}
\end{figure}

\end{document}